\documentclass[10pt,twocolumn,letterpaper]{article}

\usepackage{wacv}              % To produce the CAMERA-READY version
\usepackage{lineno}

\definecolor{wacvblue}{rgb}{0.21,0.49,0.74}
\usepackage[pagebackref,breaklinks,colorlinks,allcolors=wacvblue]{hyperref}
\usepackage{multirow}
\usepackage{graphicx}
\usepackage{amsmath}
\usepackage{amssymb}
\usepackage{booktabs}

\def\wacvPaperID{574} % *** Enter the WACV Paper ID here
\def\confName{WACV}
\def\confYear{2026}

\title{From Detection to Anticipation: Online Understanding of Struggles across Various Tasks and Activities}

\author{
Shijia Feng$^{1}$,
Michael Wray$^{1}$,
Walterio Mayol\text{-}Cuevas$^{1,2}$\\
$^{1}$University of Bristol \quad
$^{2}$Amazon\\
{\tt\small \{shijia.feng.2019, michael.wray, walterio.mayol-cuevas\}@bristol.ac.uk}
}

\begin{document}
\maketitle
\begin{abstract}
Understanding human skill performance is essential for intelligent assistive systems, with struggle recognition offering a natural cue for identifying user difficulties. While prior work focuses on offline struggle classification and localization, real-time applications require models capable of detecting and anticipating struggle online. We reformulate struggle localization as an online detection task and further extend it to anticipation—predicting struggle moments before they occur.
We adapt two off-the-shelf models as baselines for online struggle detection and anticipation. Online struggle detection achieves 70–80\% per-frame mAP, while struggle anticipation up to 2 seconds ahead yields comparable performance with slight drops. We further examine generalization across tasks and activities and analyse the impact of skill evolution. Despite larger domain gaps in activity-level generalization, models still outperform random baselines by 4–20\%.
Our feature-based models run at up to 143 FPS, and the whole pipeline, including feature extraction, operates at around 20 FPS — sufficient for real-time assistive applications.
%This runtime is adequate for real-time assistive applications.%, where immediate responses are not critical, as the recognition of struggle naturally involves a short delay, even for human observers.
\end{abstract}
    
\section{Introduction}
\label{sec:intro}
Detecting moments of struggle during human activities is essential for building intelligent assistive systems that provide timely support. Such systems include adaptive robots, wearable training devices, and elder care technologies that recognize when help is needed. In all these cases, real-time understanding of user difficulty is critical for delivering immediate feedback or intervention before failure occurs or frustration builds.

Struggle,  as defined in~\cite{feng2024strugglingdatasetbaselinesstruggle, fengevostruggle}, refers to observable difficulty in task completion, often manifested through hesitant hand movements, repeated attempts, prolonged pauses, disruptive mistakes, or expressive head and hand gestures.

Recent work~\cite{fengevostruggle} has explored offline struggle detection, also known as struggle moment localization, which identifies the start and end of struggle periods from fully observed egocentric videos. While valuable for post-hoc analysis, offline methods cannot provide real-time feedback, as they depend on complete video sequences. In contrast, online detection must work with partial observations, making it significantly more challenging due to the dynamic and variable nature of human behaviour across tasks.

Furthermore, the ability to anticipate struggle before it occurs remains largely unexplored. This limits the potential for systems to offer proactive support, not only in assistive contexts like guiding users with motor or cognitive impairments, but also for everyday users learning new skills, using complex tools, or performing high-risk tasks. Anticipating struggle could, for instance, allow an assistive collaboration robot to pre-emptively assist before failure, or enable a tutorial platform to adjust task difficulty before frustration arises. 

These gaps highlight the need for a unified framework that supports both real-time struggle detection and short-horizon anticipation, enabling timely and context-aware interventions across a range of real-world scenarios.

In this paper, we reformulate struggle moment localization as an online detection task and introduce the novel concept of struggle anticipation. This reformulation moves from offline analysis to real-time prediction, enabling proactive support but also posing new challenges, as predictions must be made from partial input.
Our study addresses two primary research questions: (1) Can struggle be detected online using partial observations? (2) Can struggle be anticipated before it occurs? To answer these, we adapt two off-the-shelf models, LSTR~\cite{xu_long_2021} and CMeRT~\cite{Pang_2025_CVPR}, as baselines for online detection and anticipation.
We also conduct in-depth experiments to investigate whether struggle patterns remain generalizable across diverse activities and tasks under these newly formulated online detection and anticipation settings. 
Unlike action recognition, which depends on task-specific motions, struggle is marked by more universal cues, such as motor hesitation or repeated attempts, making cross-task detection more feasible.
% The underlying hypothesis is that struggle manifests through common behavioural cues—such as motor hesitation, repeated attempts, or prolonged pauses—that can occur across different tasks, regardless of the specific action being performed. Unlike action recognition, which relies on task-specific motion patterns and context, struggle detection focuses on recognizing these universal indicators of difficulty, making cross-task generalization plausible.
%While prior work~\cite{fengevostruggle} explored struggle evolution and generalization in offline settings, our study extends this analysis to real-time scenarios, evaluating the persistence of struggle patterns when predictions are made from partial observations.

Our contributions are summarized as follows: 
% (1) We reformulate struggle moment localization as an online detection task, enabling real-time identification of struggle from partial observations, and further extend this to struggle anticipation, aiming to predict struggle before it occurs. (2) We provide baseline results by adapting two off-the-shelf models, LSTR~\cite{xu_long_2021} and CMeRT~\cite{Pang_2025_CVPR}, for both online struggle detection and anticipation. (3) We conduct extensive experiments to evaluate the generalizability of struggle patterns across diverse activities and tasks under online detection and anticipation settings, extending prior offline analyses~\cite{fengevostruggle}.
\begin{itemize}
    \item We reformulate struggle moment localization as an online detection task, enabling real-time identification of struggle from partial observations, and further extend this to struggle anticipation, aiming to predict struggle before it occurs.
    % \item We introduce the novel concept of struggle anticipation, addressing the proactive prediction of struggle before it occurs.
    \item We provide baseline results by adapting two off-the-shelf models, LSTR~\cite{xu_long_2021} and CMeRT~\cite{Pang_2025_CVPR}, for both online struggle detection and anticipation.
    \item We conduct extensive experiments to evaluate the generalizability of struggle patterns across diverse activities and tasks under online detection and anticipation settings. %extending prior offline analyses~\cite{fengevostruggle}.
\end{itemize}

\section{Related Work}
\label{sec:related_work}

\paragraph{Struggle vs. Mistakes/Errors.}
Understanding how people struggle during task execution has recently been explored in~\cite{feng2024strugglingdatasetbaselinesstruggle, fengevostruggle}. The study in~\cite{feng2024strugglingdatasetbaselinesstruggle} focuses on struggle recognition from trimmed video clips across three scenarios: plumbing assembly, tent pitching, and the Tower of Hanoi. Recently,~\cite{fengevostruggle} introduced the EvoStruggle dataset, comprising over 60 hours of recordings across 18 diverse tasks and emphasizing temporal localization of struggle in untrimmed videos using precise start and end annotations. While both studies contribute valuable insights into struggle determination, they are constrained to trimmed clips or offline temporal localization and do not tackle the challenges of online detection or anticipation.

Closely related areas, such as mistake or error detection~\cite{9022634, Sener_2022_CVPR, Ghoddoosian_2023_ICCV, Wang_2023_ICCV, Lee_2024_CVPR, mazzamuto2024gazingmisstepsleveragingeyegaze, Schoonbeek_2024_WACV, Flaborea_2024_CVPR}, primarily focus on procedural step recognition and identifying whether task steps are performed correctly. These datasets and approaches often annotate the mistakes/errors based on whether a task step was completed correctly. In contrast, struggle is a complementary and more nuanced concept, capturing difficulty or hesitation during task execution regardless of correctness. One may struggle without making a mistake (e.g., fumbling but succeeding), make a mistake without struggling (e.g., confidently executing the wrong step), or do both (e.g., appearing confused while failing). %Unlike mistake detection, understanding struggle requires modelling subtle behavioural cues and temporal dynamics beyond task outcomes.

\paragraph{Online Action Detection and Anticipation.} 
Online action understanding encompasses two primary tasks: online action detection and action anticipation. Online action detection focuses on classifying actions in real-time using video data up to the current frame, in contrast to offline temporal action localization, which uses the entire video. Action anticipation, by contrast, aims to predict whether a person will struggle in the next few seconds, based solely on their behaviour observed up to the current moment.

% For online action detection, recent methods include spatio-temporal localization using real-time bounding boxes across frame sequences~\cite{kopuklu2019yowo, Kim_2024_WACV} and frame-wise classification leveraging observed history, as seen in models like Long Short-Term Transformer (LSTR~\cite{xu_long_2021}), and TeSTra~\cite{zhao2022testra} that employs temporally smoothed attention for improved efficiency. 
For online action detection, recent methods fall into two main categories. Some approaches focus on spatio-temporal localization, using real-time bounding boxes across frame sequences~\cite{kopuklu2019yowo, Kim_2024_WACV}. Others perform frame-wise classification based on the observed history. Examples include models like Long Short-Term Transformer (LSTR~\cite{xu_long_2021}) and TeSTra~\cite{zhao2022testra}, which employs temporally smoothed attention to enhance efficiency.
% Unified frameworks have also emerged, such as the Memory-and-Anticipation Transformer (MAT~\cite{MAT_Wang_2023_ICCV}), which refines current predictions using future frame estimates but may introduce causal leakage when the proposed memory-anticipation circular decoder refines the memory features using the anticipation feature. 
Unified frameworks have also emerged, such as the Memory-and-Anticipation Transformer (MAT~\cite{MAT_Wang_2023_ICCV}). This MAT architecture refines current predictions using future frame estimates. However, it may introduce causal leakage when the proposed memory-anticipation circular decoder refines the memory features using the anticipation feature.
%CMeRT~\cite{Pang_2025_CVPR} addresses future information leakage by discarding the feedback circular connection from the anticipation feature and utilising only the near-future anticipation as an auxiliary task during training, and reducing training–inference discrepancies by integrating near-past and near-future context, achieving state-of-the-art performance. 
CMeRT~\cite{Pang_2025_CVPR} mitigates future information leakage by removing the circular feedback from anticipation features and using near-future prediction only as an auxiliary task during training. It also reduces training–inference mismatch by integrating near-past and near-future context to enrich the short-term memory, achieving state-of-the-art performance.

Action anticipation methods range from early recurrent models like Rolling-Unrolling LSTMs~\cite{Furnari_2019_ICCV, 9088213} and 3D CNN-based approaches~\cite{9956090} to advanced transformer-based architectures~\cite{Girdhar_2021_ICCV, Roy_2024_WACV}. Training strategies have also advanced, with DCR~\cite{Xu_2022_CVPR} using gradual frame masking and Ub-RULSTM and Ub-DCR~\cite{10510333} incorporating uncertainty modelling to enhance RULSTM and DCR performance, respectively.

\section{Problem Formulation}
Traditional struggle moment localization~\cite{fengevostruggle} relies on access to the entire video during inference, utilizing both past and future frames. In contrast, we propose a causal, real-time formulation of struggle detection, addressing two frame-level tasks: (1) \textit{online struggle detection}, which determines whether a struggle is occurring at the current frame based solely on past and present feature observations, and (2) \textit{struggle anticipation}, which predicts whether a struggle will occur at each frame within a future frame interval using the same observations. Both tasks operate on frame-level feature sequences and are supervised using cross-entropy loss, aligning with real-world assistive applications for immediate detection and proactive prediction of struggle events.

\begin{figure*}[ht]
\centering
\includegraphics[width=\linewidth]{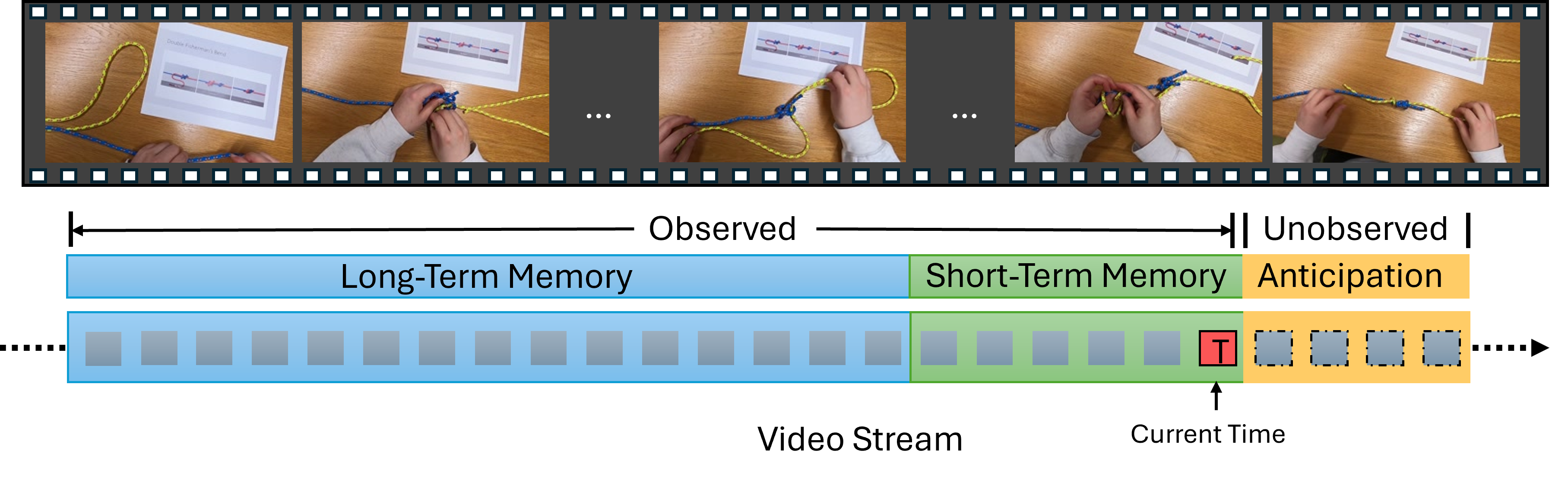}
\vspace{-5mm}
\caption{Online struggle detection and anticipation problem illustration. }
\vspace{-3mm}
\label{fig:online-detant-definition}
\end{figure*}

\subsection{Preliminaries}
Let a video sequence be denoted as $V = \{f_1, f_2, \ldots, f_N\}$, where $f_n$ is the frame at index $n$, and $N$ is the total number of frames. Each frame $f_n$ is associated with a feature vector $x_n$, extracted by a pre-trained visual encoder. The \textit{current frame} is defined as the frame at index $T$, the most recent frame available for processing. 
% At the current frame $T$, the model has access only to the sequence of features up to that frame: $X_T = \{x_1, x_2, \ldots, x_T\}$. 
At the current frame $T$, where $0 \leq T \leq N$, the model has access only to the sequence of features up to that frame: $X_T = \{x_1, x_2, \ldots, x_T\}$.

Ground-truth struggle annotations are provided as temporal intervals $\{(s_i, e_i)\}_{i=1}^M$, where $s_i$ and $e_i$ denote the start and end times of the $i$-th struggle episode. These intervals are converted to frame-level binary labels for supervision, as described in the following subsections.

\subsection{Unified Struggle Detection \& Anticipation}
% Introducing the unified framework
Both online struggle detection and anticipation aim to predict binary labels indicating the presence of struggle at or near the current frame, based on the observed feature sequence \(X_T = \{x_1, x_2, \ldots, x_T\}\), and thus can be formulated within a unified framework. 
% Both struggle online detection and anticipation tasks aim to predict binary outcomes indicating the presence of a struggle based on the observed feature sequence \(X_T = \{x_1, x_2, \ldots, x_T\}\), where \(x_n\) is the feature vector of frame \(f_n\) at index \(n\) in a video sequence \(V = \{f_1, f_2, \ldots, f_N\}\). Ground-truth struggle annotations are provided as temporal intervals \(\{(s_i, e_i)\}_{i=1}^M\), converted to frame-level binary labels for supervision.
%Online struggle detection and anticipation can be formulated within a unified framework. 
Specifically, anticipation involves predicting whether struggle will occur within the next 
$\delta$ seconds (called anticipation horizon), based on current and past observations. Detection can be viewed as a special case where the anticipation window is zero, i.e., \(\delta = 0\).
% Online struggle detection and anticipation can be unified under a single framework where detection is a special case of anticipation with an anticipation horizon \(\delta = 0\).

% Defining the unified task
In this unified framework, the goal is to predict, at the current frame \(T\), whether a struggle occurs at frame \(k \geq T\), where \(k \in [T, T+\delta]\) and \(\delta \geq 0\) is the anticipation horizon (in frames). The model outputs a binary prediction \(\hat{y}_k \in \{0, 1\}\), where \(\hat{y}_k = 1\) indicates a struggle at frame \(k\). The unified prediction function is:
\begin{equation}
\hat{y}_k = f_{\text{unified}}(X_T; \theta_{\text{unified}})
\end{equation}
for all \(k \in [T, T+\delta]\), where \(\theta_{\text{unified}}\) denotes the model parameters. Input sequences shorter than the model’s long-term memory setting are padded, and the padding positions are masked with $-\infty$ in the attention logits to exclude them from the attention computation. The ground-truth label \(y_k = 1\) if frame \(k\) falls within any struggle interval \([s_i, e_i]\) (i.e., the frame index corresponds to time \(t \in [s_i, e_i]\)), and \(y_k = 0\) otherwise.

% Specializing to detection
When \(\delta = 0\), the task reduces to online struggle detection, where the model predicts whether a struggle is occurring at the current frame \(T\). In this case, \(k = T\), and the prediction \(\hat{y}_T = f_{\text{unified}}(X_T; \theta_{\text{unified}})\) aligns with the detection objective, focusing solely on the current frame.

% Specializing to anticipation
For \(\delta > 0\), the task becomes struggle anticipation, predicting whether a struggle will occur at each frame \(k \in (T, T+\delta]\). This enables proactive interventions by forecasting future struggle events based on the current feature sequence \(X_T\).

% Training and supervision
The model is trained using cross-entropy loss to supervise the frame-level predictions \(\hat{y}_k\) for all \(k \in [T, T+\delta]\), ensuring causal processing with no access to future frames during inference. This unified approach allows a single model to handle both detection and anticipation by adjusting the anticipation horizon \(\delta\), leveraging shared feature representations and temporal dependencies in \(X_T\).

\section{Model Architecture} 
Struggle is inherently a temporal and sequential phenomenon, requiring the model to capture both spatial contexts, such as hand-object interactions, and temporal motion cues, including hand and head movements that are often reflected in background dynamics. Importantly, struggle is typically recognized over a temporal segment rather than from a single frame, highlighting the need for modelling temporal dependencies. As demonstrated in~\cite{feng2024strugglingdatasetbaselinesstruggle}, human annotators can reliably identify struggle within a 10-second window. 

Given these requirements, we first chose to adapt the LSTR model~\cite{xu_long_2021} for online struggle detection. This model, originally developed for online action detection, is particularly well-suited due to its Long Short-Term Transformer architecture. It successfully adapts the concept of RNN-based LSTMs to a transformer framework, enabling it to model exceptionally long-term dependencies, which has become a foundational component for many current transformer-based models used in online action detection and anticipation~\cite{zhao2022testra, MAT_Wang_2023_ICCV, Pang_2025_CVPR}.

As depicted in Figure~\ref{fig:online-detant-definition}, the input video feature sequence $\{x_t\}_{t=1}^N$ is partitioned into an observed segment $\{x_t\}_{t=1}^T$ and a future segment $\{x_t\}_{t=T+1}^N$, where $x_t$ denotes the video feature frame at index $t$ and $N$ is the total number of frames. 
The LSTR model~\cite{xu_long_2021} processes $\{x_t\}_{t=1}^T$ to extract a long-term memory feature sequence $M_L \in \mathbb{R}^{l \times d}$, where $l$ is the length of the processed long-term feature sequence and $d$ is the feature dimension. A transformer decoder compresses $M_L$ into a reduced representation $M_L' \in \mathbb{R}^{n \times d}$, using a fixed-size learnable query embedding $Q_L \in \mathbb{R}^{n \times d}$ with $n \ll l$.
Subsequently, the short-term memory features $M_S \in \mathbb{R}^{m \times d}$ serve as the query in another transformer decoder, attending to the key-value pairs of $M_L'$ to produce output features. These features are passed through a classifier to predict struggle or non-struggle/background labels. A standard triangular causal mask ensures that each token can only attend to past and current tokens, preventing future information leakage from $\{x_t\}_{t=T+1}^N$. To enable struggle anticipation, we concatenate $\delta = M_F$ learnable token embeddings to $M_S$, corresponding to the number of future frames to be predicted, following the modification guidance in~\cite{xu_long_2021}.

In addition, we adapt the state-of-the-art CMeRT model~\cite{Pang_2025_CVPR}, originally designed for unified online action detection and anticipation, to address the challenge of online struggle detection and anticipation. Built upon the LSTR pipeline~\cite{xu_long_2021} with additional anticipation tokens, CMeRT~\cite{Pang_2025_CVPR} manages long-term and short-term memory similarly to LSTR~\cite{xu_long_2021} but introduces three key innovations:

\begin{enumerate}
    \item \textbf{Query Vector Construction}: CMeRT~\cite{Pang_2025_CVPR} concatenates near-past memory, short-term memory $M_S \in \mathbb{R}^{m \times d}$, and anticipation token embeddings $\delta = M_F$ to form a transformer decoder query vector. This predicts frame-level struggle in short-term memory and the anticipation span $\delta$. Features corresponding to near-past memory are discarded to maintain the feature sequence length ($m + \delta$) for the initial prediction of the struggle.
    \item \textbf{Near-Future Feature Prediction}: Using compressed long-term memory $M_L' \in \mathbb{R}^{n \times d}$, CMeRT~\cite{Pang_2025_CVPR} predicts near-future features $M_F' \in \mathbb{R}^{f \times d}$ via a transformer decoder, starting from the end of the long-term memory. During training, these features are passed through a classifier to predict frame-level struggle and supervised by the cross-entropy loss; during inference, this step is skipped. Only the predicted features $M_F'$ are used in the refinement module.
    \item \textbf{Refinement Module}: A transformer decoder-based refinement module processes concatenated inputs, including compressed long-term memory $M_L'$, predicted near-future features $M_F'$, short-term memory $M_S$, and anticipation token embeddings $\delta = M_F$, as key-value pairs and uses the initial prediction of struggle as query vector to produce the final frame-level struggle prediction.
\end{enumerate}

Overall, during the training stage, all frame predictions within the short-term memory are passed through a shared classifier and supervised using the cross-entropy loss. During inference, only the prediction for the most recent frame, corresponding to the current time, is used as the output for online struggle detection.

For the classification head, we adapt both models to a binary setting, distinguishing between struggle and non-struggle, where non-struggle is treated as the background class.

\section{Experiments}
\label{sec:experiments}
This section introduces the dataset, evaluation metrics, and implementation details. We report the main results of training and evaluating the models using the same activity, also including ablations on anticipation time, and evaluate generalization across activities, tasks, and user attempts. We also provide runtime analysis and qualitative examples of model predictions on video streams.

\subsection{Dataset and Metrics}
\paragraph{Dataset}
We use the EvoStruggle dataset~\cite{fengevostruggle}, which comprises 2,793 video recordings and 5,385 annotated struggle segments with precise timestamps. It spans 18 tasks across four activity categories—Tying Knots, Origami, Tangram, and Shuffle Cards—performed five times each by 76 participants to capture skill progression. The dataset includes predefined splits for studying generalization at both the activity and task levels. For activity-level generalization, models are trained on one or more activities and evaluated on a held-out activity. For task-level generalization, models are trained on all but one task within an activity, with the excluded task used for testing.

\paragraph{Evaluation Metrics}
We report per-frame calibrated average precision (cAP)~\cite{10.1007/978-3-319-46454-1_17} for both online detection and anticipation. Unlike mean average precision (mAP)~\cite{xu_long_2021, Pang_2025_CVPR}, cAP is robust to class imbalance, with a random baseline around 50\%. This is crucial for dealing with the imbalance of the EvoStruggle dataset~\cite{fengevostruggle}, where struggle proportions vary: 0.29 (Shuffle Cards), 0.49 (Tangram), 0.25 (Origami), and 0.42 (Tying Knots). 
For anticipation, we adopt a 2-second window and use the same metric for consistency.

\subsection{Main Results}
We present the main results in this section for models trained and evaluated on each of the four activities, Tying Knots, Origami, Tangram, and Shuffle Cards, covering all tasks within each activity. This setting is referred to as within-activity evaluation. The experiments are conducted under two settings: in the first setting, models are trained and evaluated separately on each of the four activities (Tying Knots, Origami, Tangram, and Shuffle Cards), as shown in the first two columns of the table; in the second setting, models are trained on a combined set of all four activities and evaluated both on the combined data and on each activity individually, shown in the last two columns. A random baseline is also included, obtained by randomly assigning frame-level struggle labels and calculating the corresponding cAP, serving as a reference point for performance comparison. 

\begin{table}[ht]
% \scriptsize
\centering
\resizebox{0.45\textwidth}{!}{%
\begin{tabular}{ll|cccc}
\toprule
 & & \multicolumn{2}{c}{\textbf{Individual Training}} & \multicolumn{2}{c}{\textbf{Combined Training}} \\
\cmidrule(lr){3-4} \cmidrule(lr){5-6}
\textbf{Activity} & \textbf{Model} & {\textbf{Ant. (\%)} } & {\textbf{Det. (\%)} } & {\textbf{Ant. (\%)} } & {\textbf{Det. (\%)} } \\
\midrule
\multirow{3}{*}{Tying Knots}
    & Random & 51.72 & 53.54 & 51.72 & 53.54 \\
    & LSTR~\cite{xu_long_2021}   & 65.93 & 67.59 & 66.03 & 68.24 \\
    & CMeRT~\cite{Pang_2025_CVPR}  & \textbf{70.16} & \textbf{74.09} & \textbf{69.97} & \textbf{73.88} \\
\midrule
\multirow{3}{*}{Origami}
    & Random & 51.22 & 50.95 & 51.22 & 50.95 \\
    & LSTR~\cite{xu_long_2021}   & 76.00 & 74.85 & 74.62 & 75.60 \\
    & CMeRT~\cite{Pang_2025_CVPR}  & \textbf{80.82} & \textbf{83.29} & \textbf{78.11} & \textbf{80.48} \\
\midrule
\multirow{3}{*}{Tangram}
    & Random & 51.57 & 51.02 & 51.57 & 51.02 \\
    & LSTR~\cite{xu_long_2021}   & 71.76 & 69.84 & 69.27 & 64.54 \\
    & CMeRT~\cite{Pang_2025_CVPR}  & \textbf{72.45} & \textbf{75.94} & \textbf{72.34} & \textbf{75.16} \\
\midrule
\multirow{3}{*}{Shuffle Cards}
    & Random & 49.04 & 51.14 & 49.04 & 51.14 \\
    & LSTR~\cite{xu_long_2021}   & 64.57 & 71.91 & 60.02 & 64.30 \\
    & CMeRT~\cite{Pang_2025_CVPR}  & \textbf{69.12} & \textbf{73.85} & \textbf{64.94} & \textbf{70.57} \\
\midrule
\multirow{3}{*}{Overall}
    & Random & {-} & {-} & 50.41 & 49.23 \\
    & LSTR~\cite{xu_long_2021}   & {-} & {-} & 74.41 & 72.12 \\
    & CMeRT~\cite{Pang_2025_CVPR}  & {-} & {-} & \textbf{76.28} & \textbf{78.51} \\
\bottomrule
\end{tabular}%
}
% \vspace{-5mm}
\caption{Performance comparison of models trained on individual activities versus combined activities, evaluated by anticipation average cAP and online detection cAP}
\vspace{-3mm}
\label{tab:within-activity-eval}
\end{table}

Table~\ref{tab:within-activity-eval} presents the results for both struggle anticipation and online struggle detection, reported in terms of calibrated Average Precision (cAP), using the LSTR~\cite{xu_long_2021} and CMeRT~\cite{Pang_2025_CVPR} models. 
As shown in the table, both LSTR~\cite{xu_long_2021} and CMeRT~\cite{Pang_2025_CVPR} significantly outperform the random baseline across all activities and settings, demonstrating their ability to learn meaningful temporal struggle patterns. Comparing the two models, CMeRT~\cite{Pang_2025_CVPR} consistently achieves the best performance across both struggle anticipation and online struggle detection tasks. 
For example, in the Origami activity, CMeRT~\cite{Pang_2025_CVPR} reaches 80.82\% anticipation and 83.29\% online detection cAP under individual training, outperforming LSTR~\cite{xu_long_2021} by 4.82\% and 8.44\%, respectively.
A similar trend is observed in the Tying Knots activity. For the Tangram and Shuffle Cards activities, while CMeRT~\cite{Pang_2025_CVPR} still outperforms LSTR~\cite{xu_long_2021}, the performance gains are smaller, specifically in the anticipation task for Tangram and online detection for Shuffle Cards.

In terms of task comparison, the models generally achieve slightly higher or comparable performance on online struggle detection compared to struggle anticipation. This trend is particularly evident for CMeRT~\cite{Pang_2025_CVPR}, which consistently yields higher cAP scores in online detection than in anticipation across all activities. This trend is reasonable, as action anticipation is generally more challenging than online action detection, which can be even more challenging for detecting struggle and anticipating whether a person will struggle based on observations up to the current time.  Predicting future frames is inherently difficult due to the potential for sudden changes in human behaviour—actions can shift dramatically, and goals may change unexpectedly (e.g., a person may abruptly switch tasks). 

Regarding the training strategy, models trained on the combined dataset generally achieve comparable performance to those trained individually, with only a slight drop in cAP. This minor decline may be attributed to the increased difficulty of learning to detect struggle across a broader range of diverse activities, which can introduce additional variability and distract the model.

The experiment results with additional event-level evaluation metrics are included in Supp. Section 3. 

% To fairly compare LSTR~\cite{xu_long_2021} and CMeRT~\cite{Pang_2025_CVPR}, precision-recall (PR) curves are plotted across all activities for both online struggle detection and anticipation, offering a more comprehensive evaluation, especially suited to the imbalanced nature of struggle data.
% Figure~\ref{fig:online-det-PR-model-comparison} presents the PR curves for online detection, while Figure~\ref{fig:online-ant-PR-model-comparison} shows the PR curves for anticipation. The Area Under the Curve (AUC) is also computed for each PR curve to quantify performance. 
% As shown in the two figures, the PR curves for CMeRT~\cite{Pang_2025_CVPR} generally exhibit larger AUCs than those of LSTR~\cite{xu_long_2021}, with particularly notable gains in the Origami activity. The performance gap is smaller in the Tangram and Shuffle Cards activities. These results further demonstrate that CMeRT~\cite{Pang_2025_CVPR} is more effective at online struggle detection and anticipation.

% \begin{figure}[t]
% \centering
% \includegraphics[width=\linewidth]{wacv-2026-author-kit-template/figures/LSTR-cAP-anticipation-time.png}
% \caption{The impact of the struggle anticipation time length using the model LSTR~\cite{xu_long_2021}.}
% \vspace{-5mm}
% \label{fig:lstr-ant-length}
% \end{figure}

\subsection{Ablation Study on Anticipation Time}

\begin{figure}[ht]
\centering
\includegraphics[width=\linewidth]{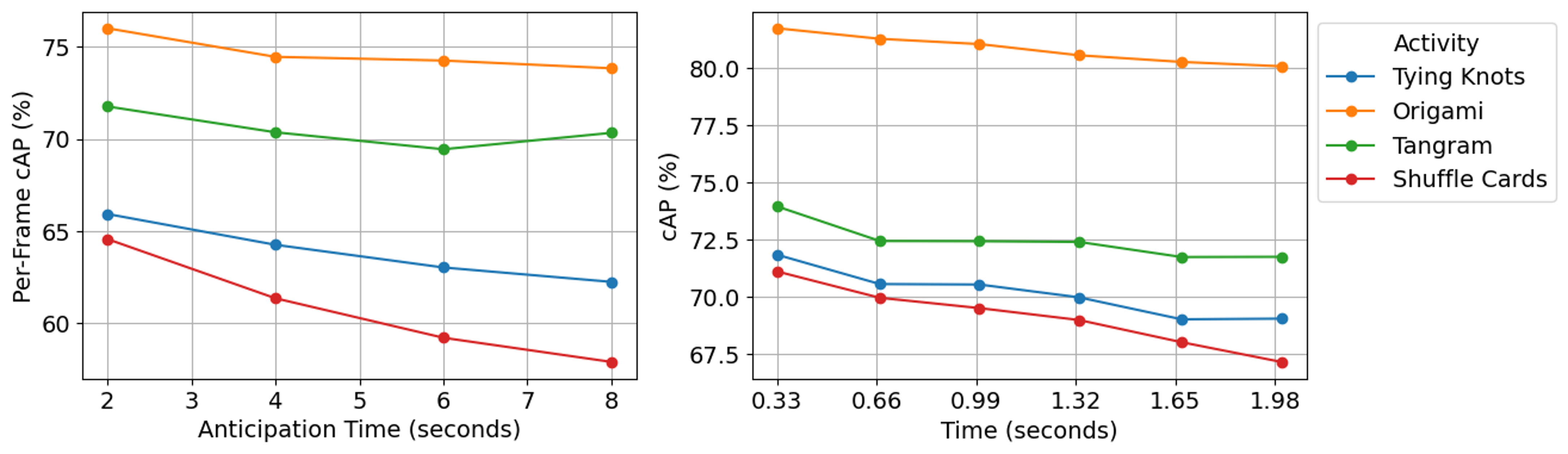}
\caption{The left plot shows the effect of anticipation time length using LSTR~\cite{xu_long_2021}, while the right shows the impact of anticipation interval with CMeRT~\cite{Pang_2025_CVPR}.}
\vspace{-3mm}
\label{fig:anticipation-time-ablation}
\end{figure}

For the struggle anticipation task, different choices of anticipation time length are further investigated as an ablation study. 
We first train the LSTR model~\cite{xu_long_2021} with varying anticipation lengths ranging from 2 to 8 seconds, in 2-second intervals. For each activity, we plot the corresponding cAP scores, as shown in the left plot in Figure~\ref{fig:anticipation-time-ablation}. The overall trend shows that cAP scores decline as the anticipation length increases, which is consistent with the intuition that longer anticipation windows are more challenging due to the inherent uncertainty of human behaviour. This drop is most pronounced in the Shuffle Cards activity, where struggle events are often abrupt and catastrophic—participants may suddenly drop cards while shuffling. In contrast, the decline in cAP scores is less steep for the Origami and Tangram activities, likely because struggles in these tasks tend to be more gradual and prolonged, as participants work through the challenges of either task completion or puzzle solving. 

% \begin{figure}[ht]
% \centering
% \includegraphics[width=\linewidth]{wacv-2026-author-kit-template/figures/CMeRT-cAP-anticipation-interval.png}
% \caption{The impact of the struggle anticipation interval of the model CMeRT~\cite{Pang_2025_CVPR}.}
% % \vspace{-5mm}
% \label{fig:cmert-ant-interval}
% \end{figure}

We also present line plots showing the cAP scores of CMeRT~\cite{Pang_2025_CVPR} across anticipation frame intervals over a 2-second window (6 frames with around 0.33–0.34 second spacing), as shown in the right plot in Figure~\ref {fig:anticipation-time-ablation}. The cAP scores generally decrease toward later frames, maintaining the same trend even at finer temporal granularity.

\subsection{Generalization Across Activities and Tasks} 

\begin{table}[ht]
% \scriptsize
\centering
\resizebox{0.45\textwidth}{!}{%
\begin{tabular}{llcc}
\toprule
\textbf{Activity} & \textbf{Gen. Type} & {\textbf{Ant. Avg. (\%)} } & {\textbf{Det. cAP (\%)} } \\
\midrule
\multirow{2}{*}{Tying Knots}
    & Random & 51.72 & 53.54 \\
    & Activity-Level & 61.90 & 64.10 \\
\midrule
\multirow{2}{*}{Origami}
    & Random & 51.22 & 50.95 \\
    & Activity-Level & 69.70 & 70.56 \\
\midrule
\multirow{2}{*}{Tangram}
    & Random & 51.57 & 51.02 \\
    & Activity-Level & 62.58 & 63.56 \\
\midrule
\multirow{2}{*}{Shuffle Cards}
    & Random & 49.04 & 51.14 \\
    & Activity-Level & 55.02 & 55.38 \\
\bottomrule
\end{tabular}%
}
% \vspace{-5mm}
\caption{CMeRT~\cite{Pang_2025_CVPR} model performance for activity-level generalization, evaluated by anticipation average and detection cAP.}
\vspace{-5mm}
\label{tab:activity-level-gen}
\end{table}

\begin{table*}[ht]
\scriptsize
\centering
% \resizebox{0.45\textwidth}{!}{%
\begin{tabular}{llcccccccccc}
\toprule
\multirow{2}{*}{\textbf{Activity}} & \multirow{2}{*}{\textbf{Gen. Type}} & \multicolumn{2}{c}{\textbf{Task 01}} & \multicolumn{2}{c}{\textbf{Task 02}} & \multicolumn{2}{c}{\textbf{Task 03}} & \multicolumn{2}{c}{\textbf{Task 04}} & \multicolumn{2}{c}{\textbf{Task 05}} \\
\cmidrule(lr){3-4} \cmidrule(lr){5-6} \cmidrule(lr){7-8} \cmidrule(lr){9-10} \cmidrule(lr){11-12}
& & {\textbf{Ant.}} & {\textbf{Det.}} & {\textbf{Ant.}} & {\textbf{Det.}} & {\textbf{Ant.}} & {\textbf{Det.}} & {\textbf{Ant.}} & {\textbf{Det.}} & {\textbf{Ant.}} & {\textbf{Det.}} \\
\midrule
\multirow{2}{*}{Tying Knots}
    & Random & 52.62 & 52.22 & 52.02 & 52.50 & 52.40 & 51.50 & 50.91 & 52.31 & 51.04 & 48.88 \\
    & Task-Level & 70.18 & 74.52 & 70.65 & 74.28 & 70.30 & 74.09 & 67.91 & 71.06 & 72.82 & 74.91 \\
\midrule
\multirow{2}{*}{Origami}
    & Random & 50.36 & 49.33 & 48.07 & 48.28 & 50.20 & 49.47 & 51.34 & 47.65 & {-} & {-} \\
    & Task-Level & 74.46 & 75.91 & 81.53 & 83.61 & 80.22 & 82.45 & 77.80 & 79.30 & {-} & {-} \\
\midrule
\multirow{2}{*}{Tangram}
    & Random & 49.44 & 49.40 & 49.86 & 49.82 & 49.43 & 49.25 & 50.55 & 50.28 & {-} & {-} \\
    & Task-Level & 65.56 & 68.65 & 67.79 & 70.06 & 72.57 & 74.65 & 70.85 & 74.12 & {-} & {-} \\
\midrule
\multirow{2}{*}{Shuffle Cards}
    & Random & 50.35 & 50.55 & 49.12 & 49.31 & 48.77 & 49.86 & 48.51 & 49.06 & 49.27 & 50.03 \\
    & Task-Level & 56.64 & 60.04 & 64.03 & 69.02 & 54.76 & 60.49 & 57.84 & 64.26 & 69.88 & 74.38 \\
\bottomrule
\end{tabular}
% }
\caption{Performance of task-level generalization across tasks, evaluated by Anticipation Average (Ant.) and Detection cAP (Det.).}
\vspace{-5mm}
\label{tab:task-level-generalization}
\end{table*}

Based on the main results from the within-activity evaluations, we identify CMeRT~\cite{Pang_2025_CVPR} as the strongest model for both online struggle detection and anticipation. Additionally, we observe a general decline in cAP scores as the anticipation length increases in the ablation study. To maintain consistency, we fix the anticipation length at 2 seconds for subsequent experiments and further investigate the model's generalizability across activities and tasks using CMeRT~\cite{Pang_2025_CVPR}.

We explore generalization along two dimensions: (1) \textbf{Activity-level generalization}, where the model is trained on a combination of three activities and evaluated on a held-out, unseen activity—thus lacking prior knowledge of that activity; and (2) \textbf{Task-level generalization}, where the model is trained on all but one task and evaluated on the held-out, unseen task within each activity—indicating a lack of task-specific knowledge. 

% \begin{figure}[ht]
% \centering
% \includegraphics[width=0.9\linewidth]{wacv-2026-author-kit-template/figures/activity-gen-5by5-matrix-onlinedet.png}
% \caption{Heatmaps showing cross-activity generalization evaluation for \textit{online struggle detection} in a zero-shot setting.}
% % \vspace{-5mm}
% \label{fig:activity-gen-heatmap-onlinedet}
% \end{figure}

% \begin{figure}[ht]
% \centering
% \includegraphics[width=0.9\linewidth]{wacv-2026-author-kit-template/figures/activity-gen-5by5-matrix-anticipation.png}
% \caption{Heatmaps showing cross-activity generalization evaluation for \textit{struggle anticipation} in a zero-shot setting.}
% % \vspace{-5mm}
% \label{fig:activity-gen-heatmap-anticipation}
% \end{figure}

The activity-level generalization experiment results are shown in Table~\ref{tab:activity-level-gen}. The reported cAP scores consistently outperform the random baseline, indicating that the model has learned generalizable patterns across activities, even without direct exposure to the evaluated activity during training. Among all activities, Origami shows the highest improvement over the baseline—18.48\% for struggle anticipation and 19.61\% for online struggle detection, suggesting that struggles in Origami are the most transferable from other activities. In contrast, Shuffle Cards yields the smallest gains—only 5.98\% for anticipation and 4.24\% for online detection—implying that struggles in this activity are harder to generalize. This may be due to the distinct nature of Shuffle Cards, where actions like card shuffling differ significantly from the task-oriented processes in Origami, Tying Knots, and Tangram. 

\begin{figure}[ht]
\centering
\includegraphics[width=\linewidth]{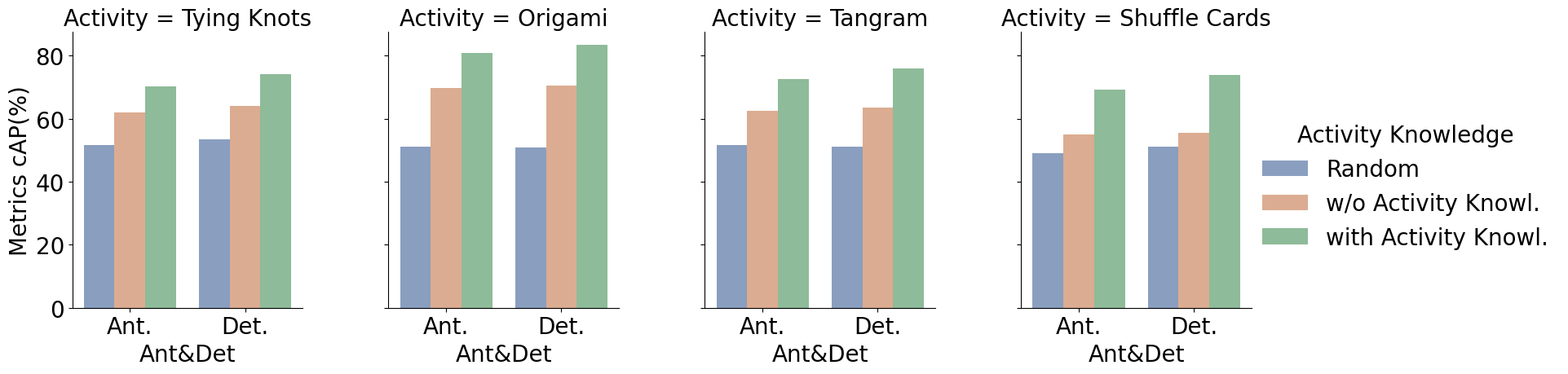}
\caption{Online struggle detection and anticipation generalization with/without Activity Knowledge.}
\vspace{-5mm}
\label{fig:antdet-comparison-activity-knowledge}
\end{figure}

We also compare the activity-level generalization results with the within-activity evaluation for each activity, on both online struggle detection and anticipation, to illustrate the impact of activity-specific knowledge. As shown in Figure~\ref{fig:antdet-comparison-activity-knowledge}, the cAP scores from models trained with activity knowledge (within-activity evaluation) are generally higher than those without it (activity-level generalization), and both outperform the random baseline. This highlights the model’s generalization ability while confirming the benefit of activity-specific training data.
% This highlights the effectiveness of the model’s generalization ability while confirming the benefit of having activity-specific training data.

\begin{figure}[ht]
\centering
\includegraphics[width=1.0\linewidth]{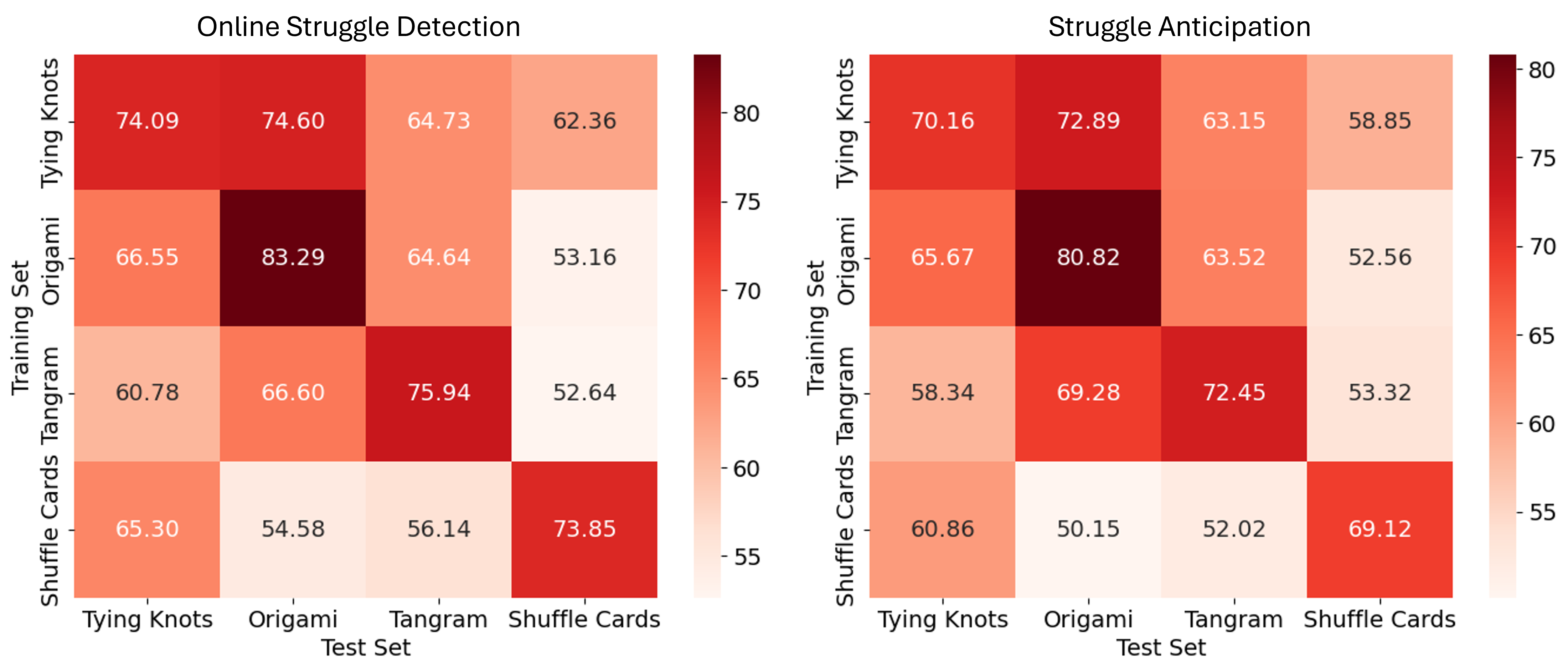}
\caption{Heatmaps showing cross-activity generalization evaluation for \textit{online struggle detection} (left) and \textit{struggle anticipation} (right) in a zero-shot setting.}
\vspace{-5mm}
\label{fig:activity-gen-heatmap-combined}
\end{figure}

We assess cross-activity generalization via zero-shot evaluation, using models trained on each activity (from within-activity experiments) to evaluate on the other three unseen activities. Figure~\ref{fig:activity-gen-heatmap-combined} shows cAP heatmaps: online detection (left) and anticipation (right). Diagonal blocks (within-activity) yield the highest scores, while off-diagonal blocks reflect zero-shot performance. Tying Knots generalizes best and transfers well to Origami. Shuffle Cards is the most challenging--models trained on other activities perform poorly on it, and vice versa. %These patterns align with Table~\ref{tab:activity-level-gen}.
% Another way to assess the model's ability to generalize across activities is through zero-shot evaluation on unseen activities. Specifically, we use the models trained on each individual activity (as obtained from the within-activity evaluation experiments) and evaluate them on the validation sets of the other three unseen activities. Figure~\ref{fig:activity-gen-heatmap-combined} presents the heatmaps of the zero-shot evaluation results for online struggle detection on the left side, while the corresponding heatmaps for struggle anticipation are shown on the right side. The diagonal blocks represent within-activity evaluations and thus yield the highest cAP scores, as there is no activity gap. The off-diagonal blocks correspond to cross-activity evaluations, where the model is further evaluated on activities it was not trained on in a zero-shot setting. Among the activities, Tying Knots appears to generalize best from other activities, followed by Origami and Tangram. Notably, the model trained on Tying Knots performs particularly well when evaluated on Origami. 
% In contrast, Shuffle Cards poses the greatest challenge for generalization: models trained on the other activities perform poorly on Shuffle Cards, and the model trained on Shuffle Cards does not transfer well to the remaining activities. 
% These observations are consistent with the trends discussed in Table~\ref{tab:activity-level-gen}. 

\begin{figure}[ht]
\centering
\includegraphics[width=\linewidth]{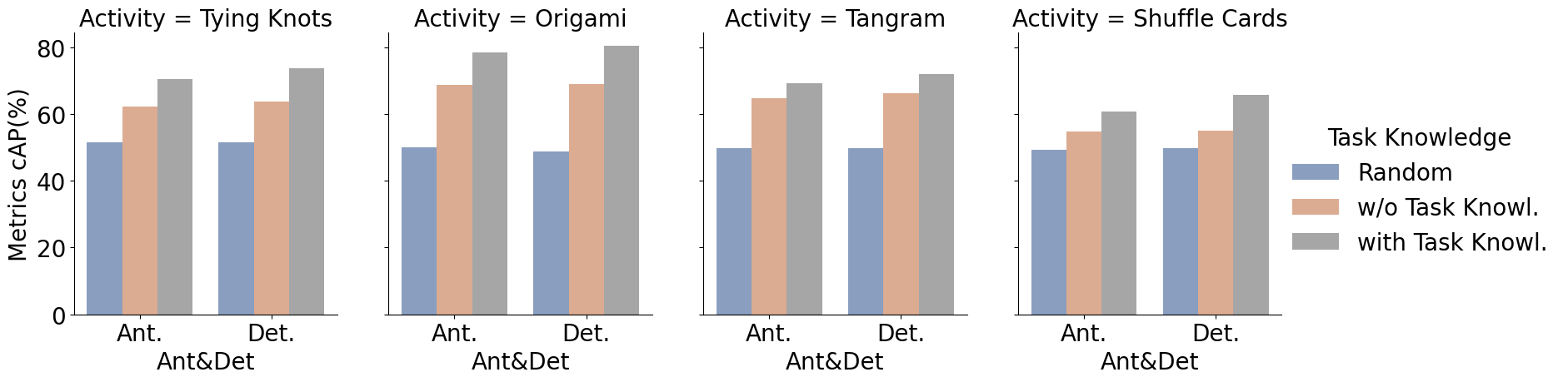}
\caption{Online struggle detection and anticipation generalization with/without Task Knowledge.}
\vspace{-3mm}
\label{fig:antdet-comparison-task-knowledge}
\end{figure}

The task-level generalization results are presented in Table~\ref{tab:task-level-generalization}. The cAP scores consistently outperform the random baseline, indicating that the model can capture the struggle patterns that are generalizable to unseen tasks within the same activity. However, task-level generalization is notably more challenging within the Shuffle Cards activity, particularly for struggle anticipation, with cAP scores dropping to 56.64\%, 54.76\%, and 57.84\% for Tasks 01, 03, and 04, respectively. This may be due to the greater variability among tasks in Shuffle Cards. In contrast, the model shows stronger task-level generalization performance across tasks within the Tying Knots, Origami, and Tangram activities. Besides, struggle anticipation cAP scores are generally lower than those for online struggle detection, as observed in previous experiments. 
Finally, Figure~\ref{fig:antdet-comparison-task-knowledge} highlights the importance of task-specific knowledge. The average cAP scores across the four or five tasks from the task-level generalization experiments—where the model has been trained on the other tasks within the same activity—are higher than those from the activity-level generalization setting, where the model is evaluated on tasks without any prior task-specific knowledge as the whole activity is unseen to the model. This demonstrates that having knowledge of the tasks within an activity significantly improves performance.

\subsection{Impact of Skill Evolution}
We further investigate the impact of skill evolution as proposed in~\cite{fengevostruggle}, where data was collected in five repetitions/attempts to show learning progress. In contrast, this paper conducts the experiments under the online struggle detection and anticipation settings. 

\begin{figure}[ht]
\centering
\includegraphics[width=\linewidth]{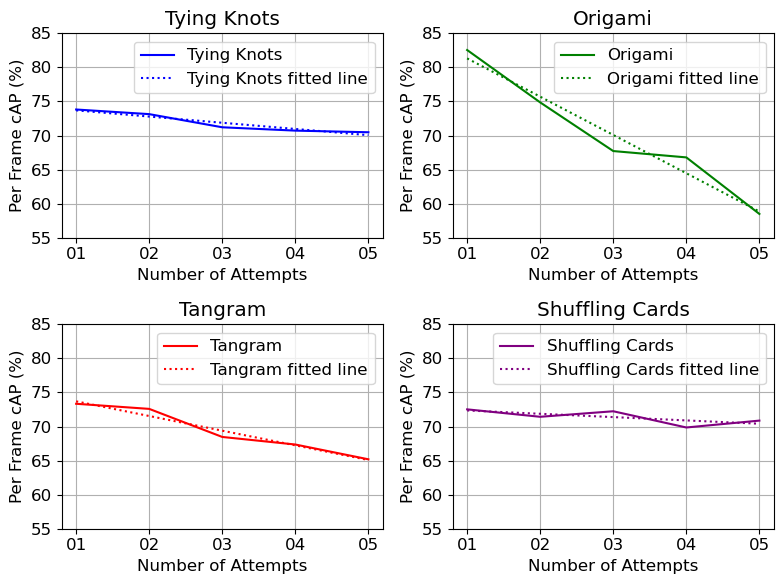}
\caption{Results of training the model using individual attempts for Online Struggle Detection.}
\vspace{-3mm}
\label{fig:evo-skill-online-det}
\end{figure}

\begin{figure}[ht]
\centering
\includegraphics[width=\linewidth]{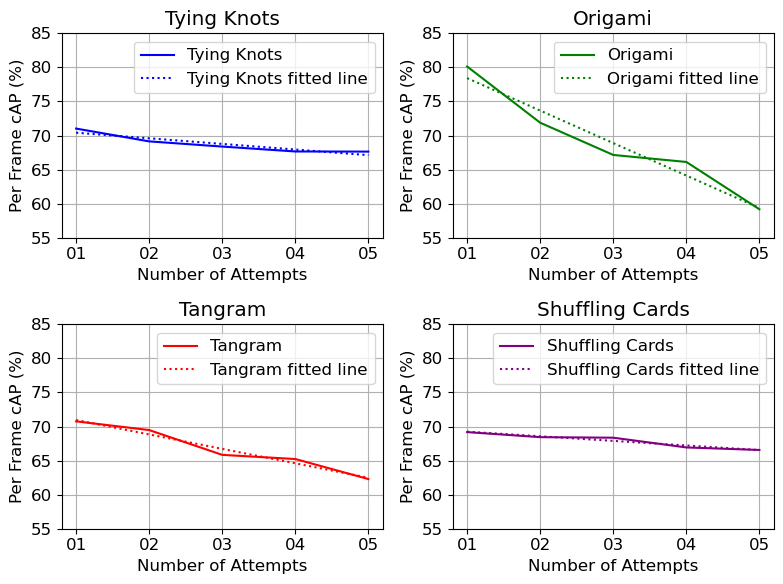}
\caption{Results of training the model using individual attempts for Struggle Anticipation.}
\vspace{-7mm}
\label{fig:evo-skill-ant}
\end{figure}

To examine how skill evolution affects struggle detection and anticipation, we train models on each attempt and evaluate them on a validation set containing all attempts. Results are shown in Figure~\ref{fig:evo-skill-online-det} (detection) and Figure~\ref{fig:evo-skill-ant} (anticipation).
We observe a consistent drop in cAP when models are trained on later attempts, across both struggle online detection and anticipation tasks. This decline is most pronounced in Origami (over 20\% drop from attempt 1 to 5), followed by Tangram ($\sim 10\%$), while Tying Knots and Shuffle Cards show smaller decreases ($\sim 5\%$).
The sharp performance drop in Origami likely stems from a rapid decline in struggle duration across repeated attempts. As participants quickly master the task, struggle becomes minimal in later trials. In contrast, Shuffle Cards involves repetitive actions that lead to relatively stable struggle durations over time. These trends reflect distinct learning curves: Shuffle Cards shows the slowest progression, Tying Knots and Tangram show steady improvement, while Origami exhibits the steepest curve, with participants gaining proficiency after only a few attempts.

\begin{figure}[ht]
\centering
\includegraphics[width=\linewidth]{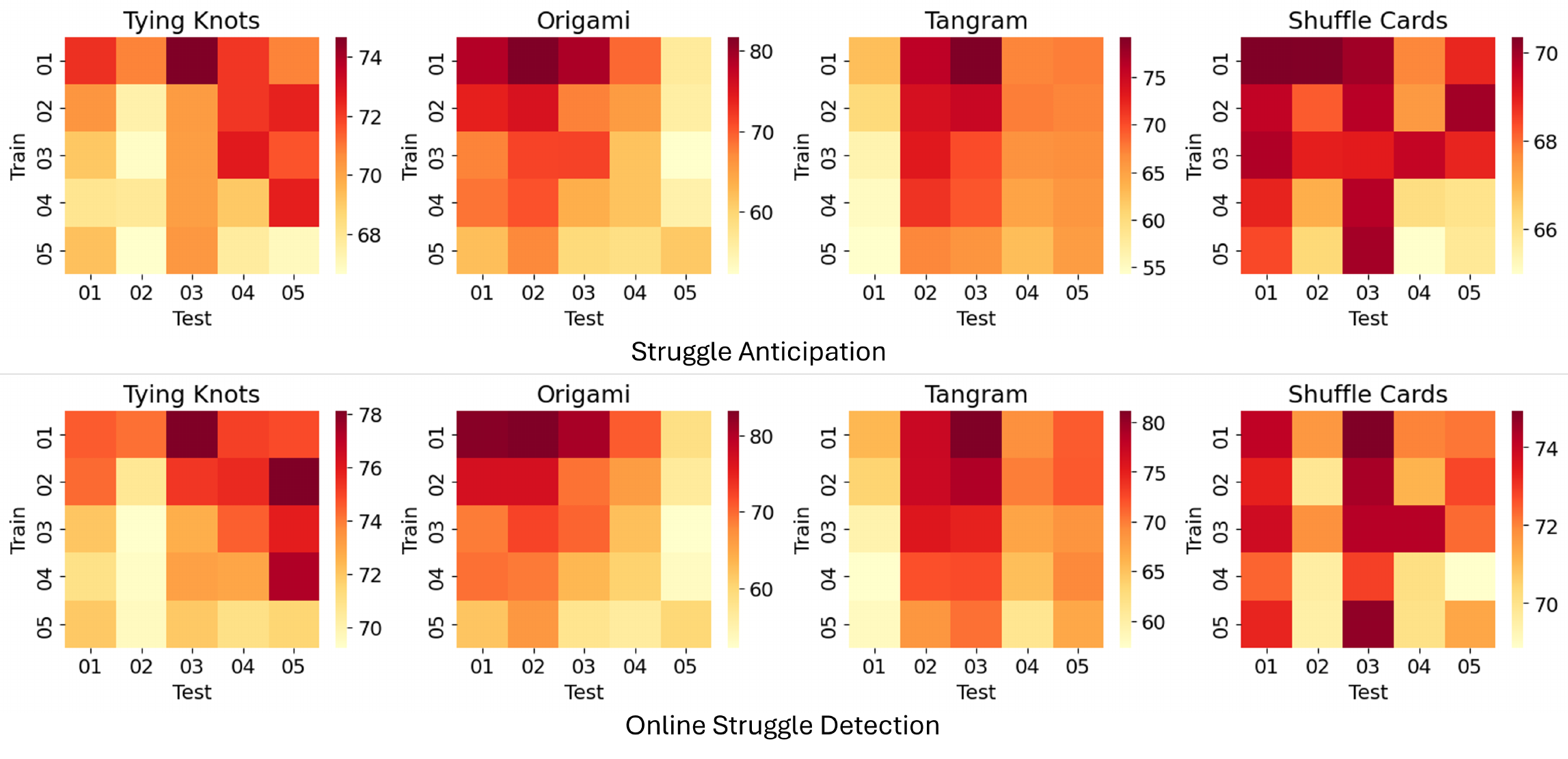}
\vspace{-6mm}
\caption{Heatmaps showing the results of training and evaluating the model on separate attempts.}
\vspace{-6mm}
\label{fig:intersection-attempts-5by5}
\end{figure}

To analyse dependencies across attempts, we train and evaluate models on all combinations of attempts, creating 5×5 matrices per activity for both online detection and anticipation. Figure~\ref{fig:intersection-attempts-5by5} shows cAP heatmaps with colour bars indicating relative performance. The top and bottom rows (detection vs. anticipation) exhibit largely consistent patterns, though variations arise across activities.
% To further examine the dependency between different training and evaluation attempts, we go beyond training models on individual attempts by also evaluating them on each separate attempt, partitioning the validation set accordingly. This results in a 5-by-5 matrix for each activity, as well as for both the online struggle detection and anticipation tasks. The results are presented in Figure~\ref{fig:intersection-attempts-5by5}, where cAP scores are visualized as heatmaps, with relative values indicated by the accompanying colour bars. Comparing the top row (online struggle detection) and the bottom row (struggle anticipation), we observe that the patterns are largely consistent across the struggle online detection and anticipation tasks, though they vary notably across different activities. 
For Tying Knots, performance gains are concentrated in the upper right of the heatmaps, showing that models trained on early attempts generalize well to later ones. In Origami, high cAP scores cluster in the upper left, indicating that early attempts are more informative, with a drop in struggle over time. For Tangram, performance peaks when evaluating on the second and third attempts, with training on earlier attempts performing slightly better, pointing to a mid-stage learning plateau. In Shuffle Cards, high scores are more evenly spread, except when both training and testing are on later attempts, reflecting a slower, steadier learning curve with less struggle variation.
% Specifically, for the Tying Knots activity, the strongest performance gains are concentrated in the upper right triangle of the heatmaps, indicating that models trained on early attempts generalize well to later attempts. This suggests a stable learning process where early-stage struggle patterns help the model anticipate later-stage behaviours. In contrast, for the Origami activity, the highest cAP scores are clustered in the upper left region, where both training and evaluation are based on early attempts. This reflects a sharp drop in struggle over time, making later attempts less useful for either training or evaluation. For the Tangram activity, performance peaks mainly when the model is evaluated on the second and third attempts, with slightly better results when trained on earlier attempts, possibly reflecting a mid-stage learning plateau. Finally, in the Shuffle Cards activity, high cAP scores are more evenly distributed, except for the models when both training and evaluation are done on later attempts, indicating a slower, more consistent learning curve with less variation in struggle over time.

\subsection{Runtime Analysis}

We evaluate the runtime of online struggle detection and anticipation using two video feature extraction backbones, SlowFast-R50~\cite{feichtenhofer2019slowfast, fan2020pyslowfast} and a lightweight backbone, S3D~\cite{xie2018rethinkingspatiotemporalfeaturelearning}, together with two temporal models, LSTR~\cite{xu_long_2021} and CMeRT~\cite{Pang_2025_CVPR}. All experiments are conducted on a single 1080Ti GPU with a batch size of 1. The results are summarised in Table~\ref{tab:runtime-analysis} and more in Supp. % The S3D provides a slightly faster backbone than SlowFast, while LSTR is significantly lighter than CMeRT. When combined, the fastest setting (S3D + LSTR) achieves 23.7 FPS, and the slowest (SlowFast + CMeRT) runs at 18.2 FPS, both of which remain feasible for real-time applications. 
We further analyse the impact of backbone and stride, shown in Table~\ref{tab:stride-backbone}.

\begin{table}[h]
\centering
\scriptsize
\begin{tabular}{lccc}
\toprule
\textbf{Model} & \textbf{GFLOPs} & \textbf{Params (M)} & \textbf{Runtime / FPS} \\
\midrule
% \multicolumn{4}{c}{\textit{Feature Extraction Backbones}} \\
% \midrule
% SlowFast-R50 & 66.42 & 33.64 & 27.80 ms / 24.6 FPS \\
% S3D          & 94.85 & 15.82 & 28.00 ms / 28.4 FPS \\
% \midrule
% \multicolumn{4}{c}{\textit{Temporal Models (feature input)}} \\
% \midrule
% LSTR         & 3.20  & 25.75 & 6.99 ms / 143 FPS \\
% CMeRT        & 5.87  & 42.58 & 14.27 ms / 70 FPS \\
% \midrule
% \multicolumn{4}{c}{\textit{Combined Runtime (Feature + Model)}} \\
% \midrule
SlowFast + LSTR  & 69.62 & 59.39 & 47.69 ms / 21.0 FPS \\
SlowFast + CMeRT & 72.29 & 76.22 & 54.97 ms / 18.2 FPS \\
S3D + LSTR       & 98.05 & 41.57 & 42.23 ms / 23.7 FPS \\
S3D + CMeRT      & 100.72 & 58.40 & 49.51 ms / 20.2 FPS \\
\bottomrule
\end{tabular}
\vspace{-2mm}
\caption{Overall runtime analysis.}
\label{tab:runtime-analysis}
\end{table}
\vspace{-6mm}

\begin{table}[h]
\centering
\scriptsize
\begin{tabular}{lcccc}
\toprule
\textbf{Setting} & \textbf{Feat. FPS} & \textbf{Ant. cAP} & \textbf{Det. cAP} \\
\midrule
SlowFast + CMeRT, stride 16 & 3.125 & 76.28 & 78.51 \\
SlowFast + CMeRT, stride 32 & 1.562 & 73.04 & 74.80 \\
S3D + CMeRT, stride 16      & 3.125 & 74.28 & 75.71 \\
\bottomrule
\end{tabular}
\vspace{-2mm}
\caption{Impact of backbone and stride on cAP (\%).}
\label{tab:stride-backbone}
\end{table}
\vspace{-5mm}

\subsection{Qualitative Results}
\begin{figure}[ht]
\centering
\includegraphics[width=\linewidth]{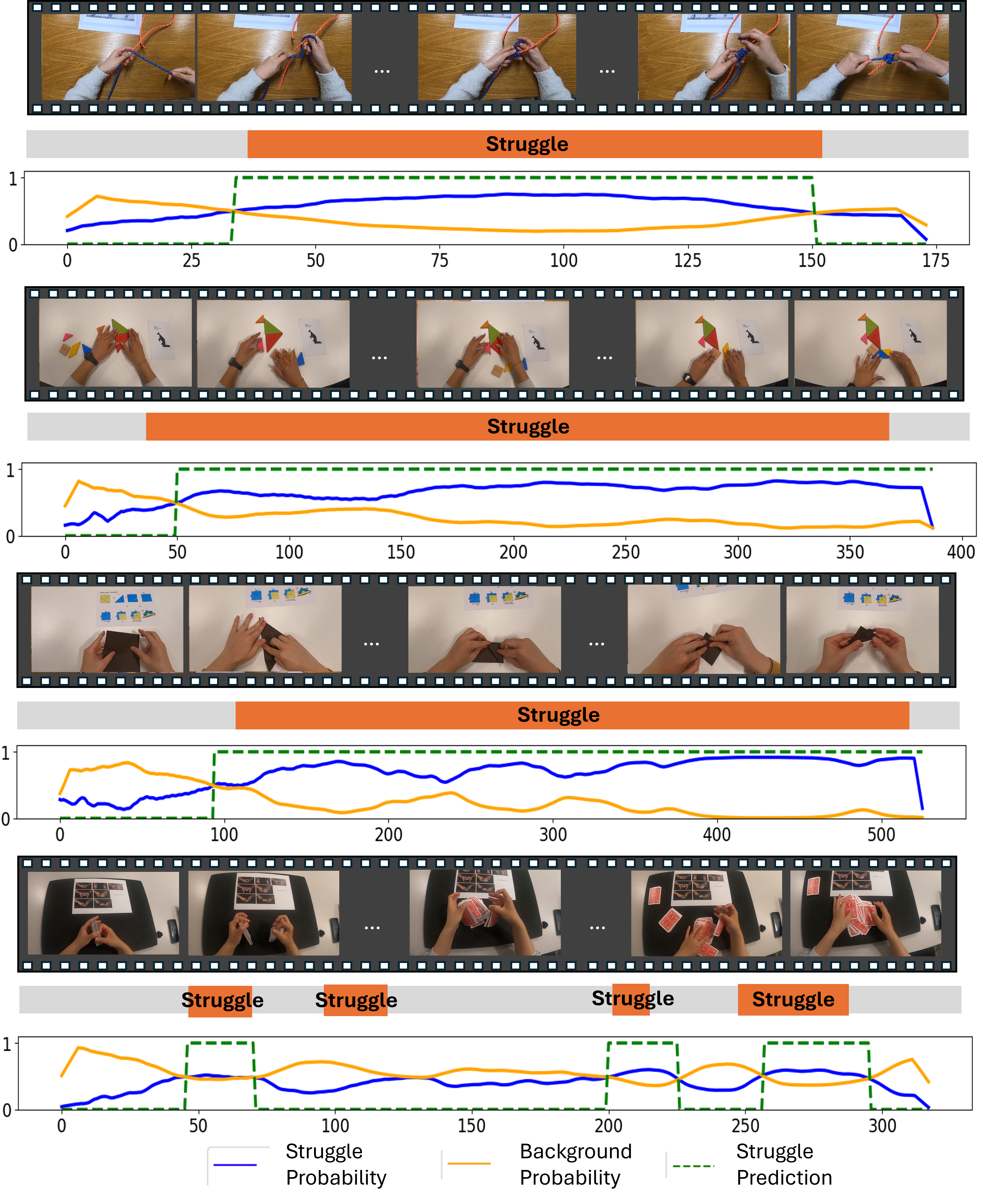}
\vspace{-7mm}
\caption{Qualitative Results on Struggle Anticipation Using Model CMeRT~\cite{Pang_2025_CVPR} Across the Four Activities (From Top to Bottom): Tying Knots, Tangram, Origami, and Shuffle Cards.}
\vspace{-8mm}
\label{fig:struggle-anticipation-visualization}
\end{figure}

The predicted probabilities of struggle and non-struggle (background) are visualized to assess model performance. As shown in Figure~\ref{fig:struggle-anticipation-visualization}, the predictions from the model CMeRT~\cite{Pang_2025_CVPR} are visualized across four activities. Model predictions are highlighted when the struggle probability exceeds that of the background. Our analysis focuses on struggle anticipation, which is generally more challenging and shares similarities with online struggle detection.

Struggle anticipation becomes especially difficult when multiple struggle moments occur in a single video. For instance, in the Shuffle Cards activity, CMeRT~\cite{Pang_2025_CVPR} produces some false positives and struggles with accurately localizing the start and end of struggle periods. In the Tying Knots, Origami, and Tangram activities, the CMeRT~\cite{Pang_2025_CVPR} model is able to anticipate struggle moments from the videos, although there may be slight shifts in the predicted time boundaries %or ambiguity when the struggle and non-struggle probabilities fluctuate around 0.5, as illustrated in the Tangram example. 
% (Struggle Anticipation Examples: (1) Tying Knots video 14_02_04 LSTR non-struggle, CMeRT detect well. (2) Tangram 05_02_03, (3) Origami 06_03_01, and (4) Shuffle Cards 02_04_02)
% different dimensions of comparison: (1) model LSTR vs. CMeRT, (2) across the four activities, and (3) online struggle detection vs. struggle anticipation--could be just using CMeRT model. 

\section{Conclusions}
\label{conclusions}
In this paper, we extend struggle detection to an online setting and explore whether struggle can be anticipated from partial video streams. Two existing models are adapted for this task, with experiments examining generalization across activities and tasks, as well as the influence of repeated attempts reflecting skill evolution.
Results show that struggle can be anticipated over time using past frames, and models can generalize to unseen domains, outperforming random baselines. Performance tends to decline when trained on later attempts, suggesting struggles become less pronounced with practice. The entire pipeline operates at approximately 20 FPS, enabling near real-time detection and anticipation. 
% Future work will involve optimizing the feature extraction process to increase speed while improving the accuracy of struggle detection. 
Future work will involve exploring the complexity of the model and the end-to-end training pipeline to achieve higher struggle detection accuracy. 

\paragraph{Acknowledgements} This work was supported jointly by the China Scholarship Council (CSC) and the University of Bristol under PhD Scholarship Award No. 202109210007. 

{
    \small
    \bibliographystyle{ieeenat_fullname}
    \bibliography{main}
}

\end{document}

\maketitle

\section{Implementation Details}
% \subsection{Implementation Details}

\begin{figure*}[ht]
\centering
\includegraphics[width=\linewidth]{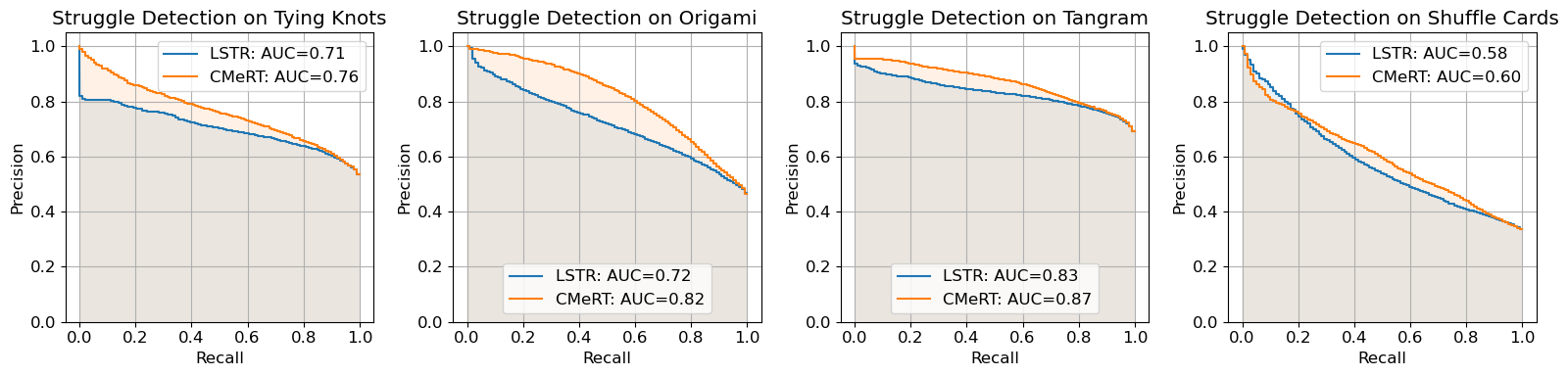}
\caption{Precision-Recall Curve of models trained on individual activities, evaluated by the cAP on the \textit{online struggle detection} task.}
% \vspace{-5mm}
\label{fig:online-det-PR-model-comparison}
\end{figure*}

\begin{figure*}[ht]
\centering
\includegraphics[width=\linewidth]{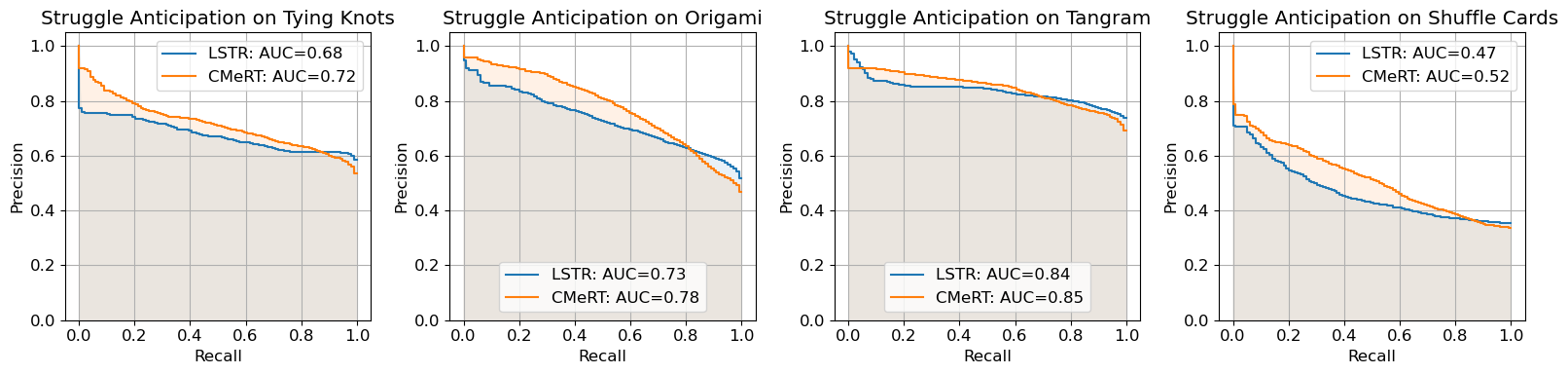}
\caption{Precision-Recall Curve of models trained on individual activities, evaluated by the cAP on the \textit{struggle anticipation} task.}
% \vspace{-5mm}
\label{fig:online-ant-PR-model-comparison}
\end{figure*}

\paragraph{Input Features.}
Since both LSTR~\cite{xu_long_2021} and CMeRT~\cite{Pang_2025_CVPR} are feature-based temporal models, we use pre-extracted video features from the EvoStruggle dataset~\cite{fengevostruggle}. These features are obtained using the SlowFast Network~\cite{feichtenhofer2019slowfast, fan2020pyslowfast}, pre-trained on the Kinetics dataset~\cite{https://doi.org/10.48550/arxiv.1705.06950}.

The SlowFast model processes each clip of 32 frames sampled from the original 50 FPS video with a stride of 16 frames, producing one feature vector per 16-frame segment. This results in a temporal resolution of one feature every $16/50 = 0.32$ seconds, corresponding to a feature-level frame rate of approximately 3.125 FPS. 

Each feature vector is 2304-dimensional and consists of two parts: a 2048-dimensional component from the Slow pathway, capturing spatial visual information, and a 256-dimensional component from the Fast pathway, encoding motion information. When loading the features, we explicitly separate these two components and feed them into the feature fusion head of the respective model. The fusion head concatenates the visual and motion features and projects them via an MLP before passing them to the temporal backbone.

For the anticipation task, the model predicts future struggle states over a target horizon (e.g., 2 seconds), corresponding to approximately 6 feature frames under a feature sampling interval of 0.32 seconds. When computing anticipation length in terms of frames, the feature frame rate is rounded to 3 FPS.

To align the ground-truth annotations with the feature frames, we convert the temporal start and end times into frame-level labels by multiplying them by the feature frame rate (3.125 FPS) and rounding to the nearest integers.

\paragraph{LSTR~\cite{xu_long_2021} for Online Detection and Anticipation.}
We train our models from scratch using the Adam optimizer with a base learning rate of $1 \times 10^{-6}$ and a weight decay of $5 \times 10^{-3}$. A cosine learning rate scheduler is applied for 20 epochs, with a linear warm-up over the first five epochs starting from $1 \times 10^{-7}$. Unless otherwise stated, the training batch size is set to 16.
For within-activity evaluations, such as the \textit{Origami} activity, we adopt the LSTR model~\cite{xu_long_2021} with the following settings:

\begin{itemize}
    \item \textbf{Architecture:} The encoder module consists of 3 transformer decoder layers for compressing long-term memory, each with a feed-forward dimension of 1024 and 16 attention heads. The decoder module comprises two transformer decoder layers that attend to the compressed long-term memory using the short-term memory as a query.
    \item \textbf{Memory Settings:} The long-term memory includes 1600 frames (equivalent to 512 seconds of video), sampled at a rate of 4. The short-term memory consists of 25 frames (8 seconds), sampled at a rate of 1.
    \item \textbf{Anticipation Settings:} For struggle anticipation, the anticipation length is set to 2 seconds, corresponding to 6 frames of input features.
    \item \textbf{Regularization:} A dropout rate of 0.2 is applied to alleviate overfitting.
\end{itemize}

The same training and architectural settings are applied to the \textit{Tangram} and \textit{Shuffle Cards} activities. For the \textit{Tying Knots} activity, the dropout rate is increased to 0.4 to alleviate overfitting.

In the experiments where the model is trained jointly on all four activities in the EvoStruggle dataset~\cite{fengevostruggle}, we extend training to 50 epochs and reduce the base learning rate to $1 \times 10^{-7}$.

\paragraph{CMeRT~\cite{Pang_2025_CVPR} for Online Detection and Anticipation.}
For within-activity evaluations, such as the \textit{Tying Knots} activity, we train the model using the Adam optimizer with a base learning rate of $5 \times 10^{-6}$ and a weight decay of $5 \times 10^{-3}$. A cosine learning rate scheduler is applied for 15 epochs, with a linear warm-up over the first five epochs starting from $5 \times 10^{-7}$. The batch size is set to 32.

\begin{itemize}
    \item \textbf{Architecture:} The CMeRT~\cite{Pang_2025_CVPR} model builds on the LSTR architecture~\cite{xu_long_2021}, with the encoder module consisting of 3 transformer decoder layers for compressing long-term memory. In addition, one transformer decoder layer is used for generating future features, and two transformer decoder layers act as the decoder module to produce the final features before classification. Each transformer layer uses eight attention heads, and the feed-forward dimension is set to 1024.
    \item \textbf{Memory Settings:} The long-term memory includes 1600 frames (512 seconds) sampled at a rate of 4. The short-term memory consists of 25 frames (8 seconds), sampled at a rate of 1.
    \item \textbf{Anticipation Settings:} The anticipation length is set to 2 seconds (6 frames), and the internal future feature generation length is set to 48 frames (16 seconds).
    \item \textbf{Regularization:} A dropout rate of 0.3 is used to alleviate overfitting.
\end{itemize}

The same training and architectural settings are applied to the \textit{Origami}, \textit{Tangram}, and \textit{Shuffle Cards} activities.

For experiments where the model is trained jointly on all four activities in the EvoStruggle dataset~\cite{fengevostruggle}, we also train for 15 epochs using the same base learning rate of $5 \times 10^{-6}$. In additional experiments, such as those examining activity-level and task-level generalization, the same hyperparameters are used.

\paragraph{Sampling Strategy for Training and Inference.}
During training, feature frames are sampled using a non-overlapping sliding window based on the short-term working memory length (25 frames, equivalent to 8 seconds). For each sampled video sequence, a random start index is selected within the video.

At inference time, we apply a batch inference strategy: the short-term memory window starts at the beginning of the video and advances one frame at a time. If there are insufficient frames preceding the short-term memory window to fill the long-term memory, we pad it by repeating the first frame index and apply masking up to the rightmost valid index. This ensures that all inputs maintain consistent dimensions while preserving causal structure.

\section{Precision-Recall Evaluation}

\begin{table*}[h]
\centering
\scriptsize
\begin{tabular}{lccccc}
\toprule
\multicolumn{6}{c}{\textbf{Struggle Anticipation}} \\
\midrule
\textbf{Activities} & \textbf{Anticipation Avg. (cAP\%)} & \textbf{ECE (frame-level)} & \textbf{Event-Level F1@$\{0.1,0.3,0.5\}$} & \textbf{ECE (event-level)} & \textbf{Lead Time (s)} \\
\midrule
Tying Knots   & 70.16 & 0.0800 & 0.5745 & 0.3763 & 1.29 \\
Origami       & 80.82 & 0.1240 & 0.3726 & 0.3726 & 2.22 \\
Tangram       & 72.45 & 0.1375 & 0.4107 & 0.3890 & 2.02 \\
Shuffle Cards & 69.12 & 0.0774 & 0.4218 & 0.3293 & 1.59 \\
\midrule
\multicolumn{6}{c}{\textbf{Online Struggle Detection}} \\
\midrule
\textbf{Activities} & \textbf{Detection (cAP\%)} & \textbf{ECE (frame-level)} & \textbf{Event-Level F1@$\{0.1,0.3,0.5\}$} & \textbf{ECE (event-level)} & \textbf{Mean Detection Delay (s)} \\
\midrule
Tying Knots   & 74.09 & 0.0909 & 0.5846 & 0.3612 & 0.68 \\
Origami       & 83.29 & 0.1376 & 0.3998 & 0.3505 & 0.01 \\
Tangram       & 75.94 & 0.1424 & 0.4574 & 0.3692 & 0.95 \\
Shuffle Cards & 73.85 & 0.0935 & 0.5269 & 0.3206 & 0.69 \\
\bottomrule
\end{tabular}
\caption{Main experimental results with additional event-level metrics, including averaged F1 scores over multiple IoU thresholds, expected calibration error (ECE), and latency measures (time-to-anticipation for struggle anticipation and mean detection delay for online struggle detection) using the CMeRT model with SlowFast features.}
\label{tab:additional-eval-metrics}
\end{table*}

To fairly compare LSTR~\cite{xu_long_2021} and CMeRT~\cite{Pang_2025_CVPR}, precision-recall (PR) curves are plotted across all activities for both online struggle detection and anticipation, offering a more comprehensive evaluation, especially suited to the imbalanced nature of struggle data.
Figure~\ref{fig:online-det-PR-model-comparison} presents the PR curves for online detection, while Figure~\ref{fig:online-ant-PR-model-comparison} shows the PR curves for anticipation. The Area Under the Curve (AUC) is also computed for each PR curve to quantify performance. 
As shown in the two figures, the PR curves for CMeRT~\cite{Pang_2025_CVPR} generally exhibit larger AUCs than those of LSTR~\cite{xu_long_2021}, with particularly notable gains in the Origami activity. The performance gap is smaller in the Tangram and Shuffle Cards activities. These results further demonstrate that CMeRT~\cite{Pang_2025_CVPR} is more effective at online struggle detection and anticipation.

\section{Additional Evaluation Metrics}
Additional evaluation metrics are reported for the main experimental results in Table~\ref{tab:additional-eval-metrics}. These include: (1) event-level averaged F1 scores at IoU thresholds of 0.1, 0.3, and 0.5, (2) expected calibration error (ECE) computed at both the frame level and the event level, and (3) latency metrics, namely time-to-anticipation (lead time) and mean detection delay. The extension experiments are conducted using the CMeRT model with SlowFast-extracted video features. The model CMeRT generates predictions at the frame level, while event-level predictions are obtained by grouping consecutive positive frames into segments defined by their start and end points. IoU is then used to determine matches between predicted struggle events and ground-truth annotations. The ECE is computed separately for frame-level and event-level outputs according to the definitions in~\cite{wang2025calibrationdeeplearningsurvey}. Anticipation lead time reflects how far in advance the model can predict the onset of a struggle, whereas mean detection delay measures the average lag in detecting a struggle in the online setting.

According to the results in Table~\ref{tab:additional-eval-metrics}, the frame-level expected calibration error (ECE) remains low (approximately 0.1), suggesting that the frame-level predictions are well calibrated. The event-level F1 scores range from 0.3726 to 0.5846, with only minor differences between the anticipation and online detection tasks. In contrast with the frame-level ECE, the event-level ECE is higher, between 0.3 and 0.4, likely due to the conversion of frame-level predictions into continuous struggle segments. For anticipation, the model achieves an average lead time of 1–2 seconds before the ground-truth struggle onset, demonstrating its ability to predict struggle in advance. Meanwhile, the mean detection delay remains within 1 second, indicating that the model responds promptly to struggle moments in the online streaming setting. 

\section{Additional Details for the Runtime Analysis}

More detailed tables and discussions are presented in this section. The full runtime results are presented in Table~\ref{full-tab:runtime-analysis}. 

Overall, the S3D provides a slightly faster backbone than SlowFast, while LSTR is significantly lighter than CMeRT. When combined, the fastest setting (S3D + LSTR) achieves 23.7 FPS, and the slowest (SlowFast + CMeRT) runs at 18.2 FPS, both of which remain feasible for real-time applications. 

\begin{table}[h]
\centering
\scriptsize
\begin{tabular}{lccc}
\toprule
\textbf{Model} & \textbf{GFLOPs} & \textbf{Params (M)} & \textbf{Runtime / FPS} \\
\midrule
\multicolumn{4}{c}{\textit{Feature Extraction Backbones}} \\
\midrule
SlowFast-R50 & 66.42 & 33.64 & 27.80 ms / 24.6 FPS \\
S3D          & 94.85 & 15.82 & 28.00 ms / 28.4 FPS \\
\midrule
\multicolumn{4}{c}{\textit{Temporal Models (feature input)}} \\
\midrule
LSTR         & 3.20  & 25.75 & 6.99 ms / 143 FPS \\
CMeRT        & 5.87  & 42.58 & 14.27 ms / 70 FPS \\
\midrule
\multicolumn{4}{c}{\textit{Combined Runtime (Feature + Model)}} \\
\midrule
SlowFast + LSTR  & 69.62 & 59.39 & 47.69 ms / 21.0 FPS \\
SlowFast + CMeRT & 72.29 & 76.22 & 54.97 ms / 18.2 FPS \\
S3D + LSTR       & 98.05 & 41.57 & 42.23 ms / 23.7 FPS \\
S3D + CMeRT      & 100.72 & 58.40 & 49.51 ms / 20.2 FPS \\
\bottomrule
\end{tabular}
% \vspace{-2mm}
\caption{Runtime analysis of feature extraction backbones and temporal models on a single NVIDIA 1080Ti GPU (batch size = 1). FPS is reported with preprocessing included.}
\label{full-tab:runtime-analysis}
\end{table}
% \vspace{-5mm}

\begin{table}[h]
\centering
\scriptsize
\begin{tabular}{lcccc}
\toprule
\textbf{Setting} & \textbf{Feat. FPS} & \textbf{Ant. cAP} & \textbf{Det. cAP} \\
\midrule
SlowFast + CMeRT, stride 16 & 3.125 & 76.28 & 78.51 \\
SlowFast + CMeRT, stride 32 & 1.562 & 73.04 & 74.80 \\
S3D + CMeRT, stride 16      & 3.125 & 74.28 & 75.71 \\
\bottomrule
\end{tabular}
% \vspace{-2mm}
\caption{Impact of backbone and stride (window size = 32) on feature throughput and cAP (\%).}
\label{supp-tab:stride-backbone}
\end{table}
% \vspace{-5mm}

We further examine the effect of backbone choice and feature stride on performance using CMeRT as the temporal model, shown in Table~\ref{supp-tab:stride-backbone}. With SlowFast at stride 16, the system processes 3.125 feature FPS and achieves 76.28\% cAP for anticipation and 78.51\% for online detection. Increasing the stride to 32 (non-overlapping windows) reduces the throughput to 1.562 feature FPS and lowers performance to 73.04\% and 74.80\%, respectively. Using S3D with a stride of 16 yields a similar 3.125 feature FPS but slightly lower accuracy (74.28\% for anticipation and 75.71\% for detection). All of the above show the following trend: A larger stride improves computational efficiency but comes with a modest drop in accuracy. S3D provides higher efficiency than SlowFast, with slightly lower accuracy overall, yet still outperforms SlowFast when a larger stride is used. These findings underscore the trade-off between backbone choice, stride selection, and model accuracy.

\begin{figure}[ht]
\centering
\includegraphics[width=\linewidth]{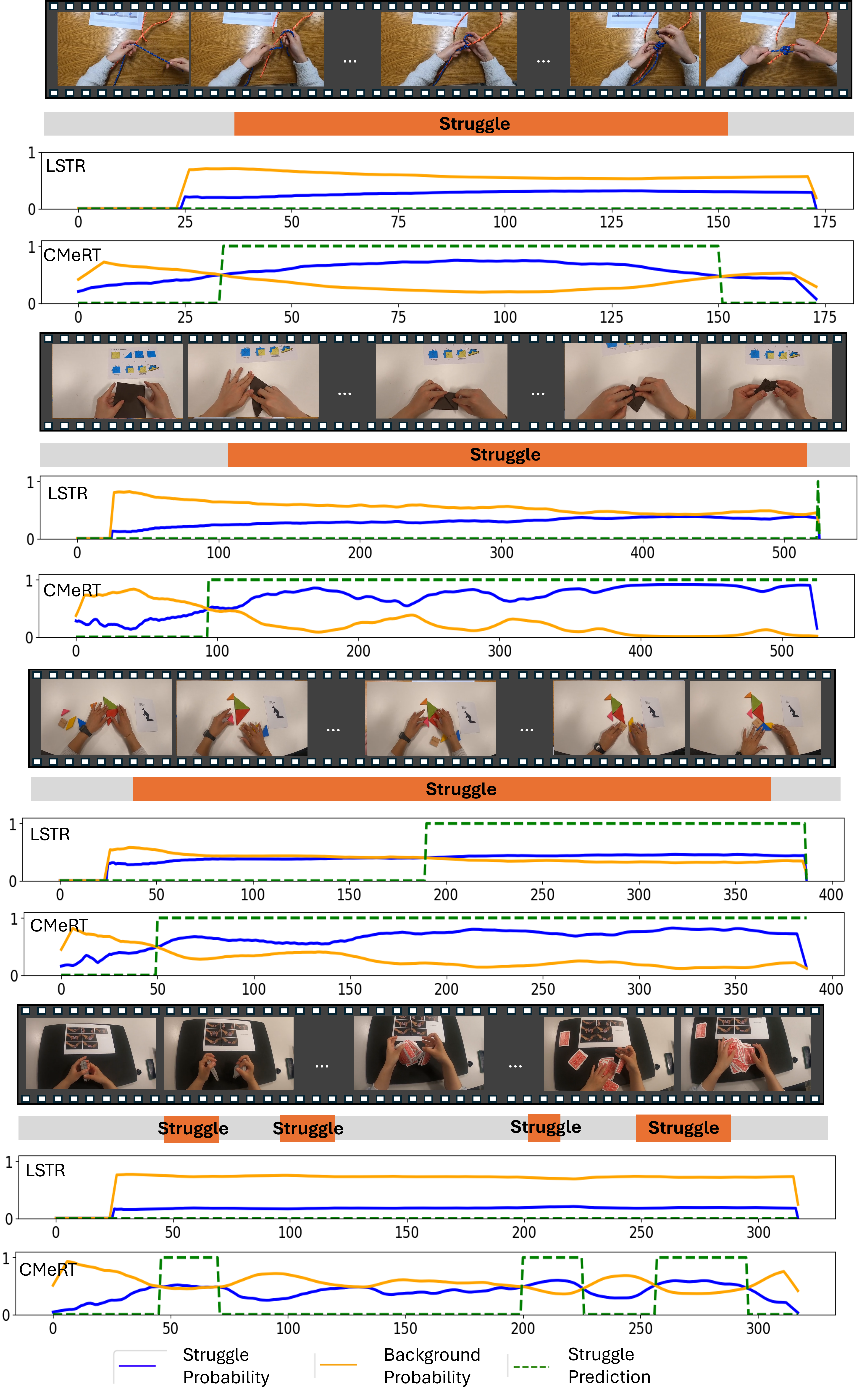}
% \vspace{-5mm}
\caption{Qualitative Results on \textbf{Struggle Anticipation} Comparing Model LSTR~\cite{xu_long_2021} and Model CMeRT~\cite{Pang_2025_CVPR} Across Activities (from top to bottom) Tying Knots, Origami, Tangram, and Shuffle Cards.}
% \vspace{-5mm}
\label{fig:struggle-anticipation-visualization-add}
\end{figure}

\section{Additional Visualization Results}

\begin{figure}[ht]
\centering
\includegraphics[width=\linewidth]{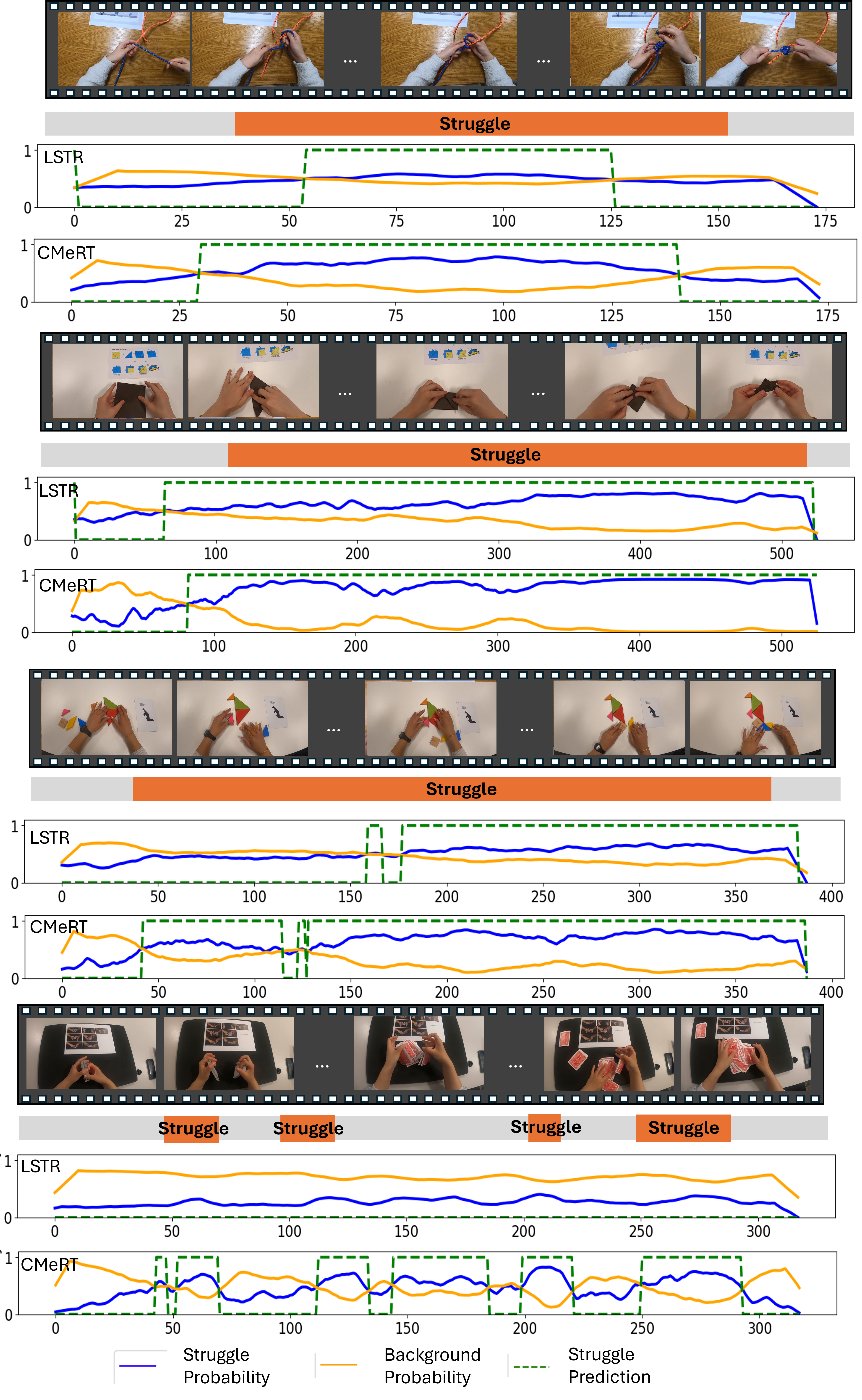}
% \vspace{-5mm}
\caption{Qualitative Results on \textbf{Online Struggle Detection} Comparing Model LSTR~\cite{xu_long_2021} and Model CMeRT~\cite{Pang_2025_CVPR} Across Activities (from top to bottom) Tying Knots, Origami, Tangram, and Shuffle Cards.}
% \vspace{-5mm}
\label{fig:struggle-online-detection-visualization-add}
\end{figure}

Figure~\ref{fig:struggle-anticipation-visualization-add} and Figure~\ref{fig:struggle-online-detection-visualization-add} present additional prediction results for struggle anticipation and online struggle detection, respectively. These visualizations include comparisons between the two models: LSTR~\cite{xu_long_2021} and CMeRT~\cite{Pang_2025_CVPR}.  

The LSTR model~\cite{xu_long_2021} is generally less accurate than CMeRT~\cite{Pang_2025_CVPR} in both detecting and anticipating struggle, partly due to the latter's use of future-feature prediction during training. While LSTR can detect some struggle moments in the online detection task, its temporal boundaries are often imprecise. In the more challenging anticipation task, LSTR frequently fails to detect any struggle moments, as illustrated in Figure~\ref{fig:struggle-anticipation-visualization-add}.

In contrast, CMeRT~\cite{Pang_2025_CVPR} demonstrates significantly better performance in both tasks, with more accurate boundaries that align closely with the ground truth. However, it still struggles with repeated or short-duration struggle moments—leading to false negatives in anticipation (e.g., Shuffle Cards activity in Figure~\ref{fig:struggle-anticipation-visualization-add}) and occasional false positives in detection (Figure~\ref{fig:struggle-online-detection-visualization-add}), highlighting the inherent difficulty of both tasks.

\begin{figure}[ht]
\centering
\includegraphics[width=\linewidth]{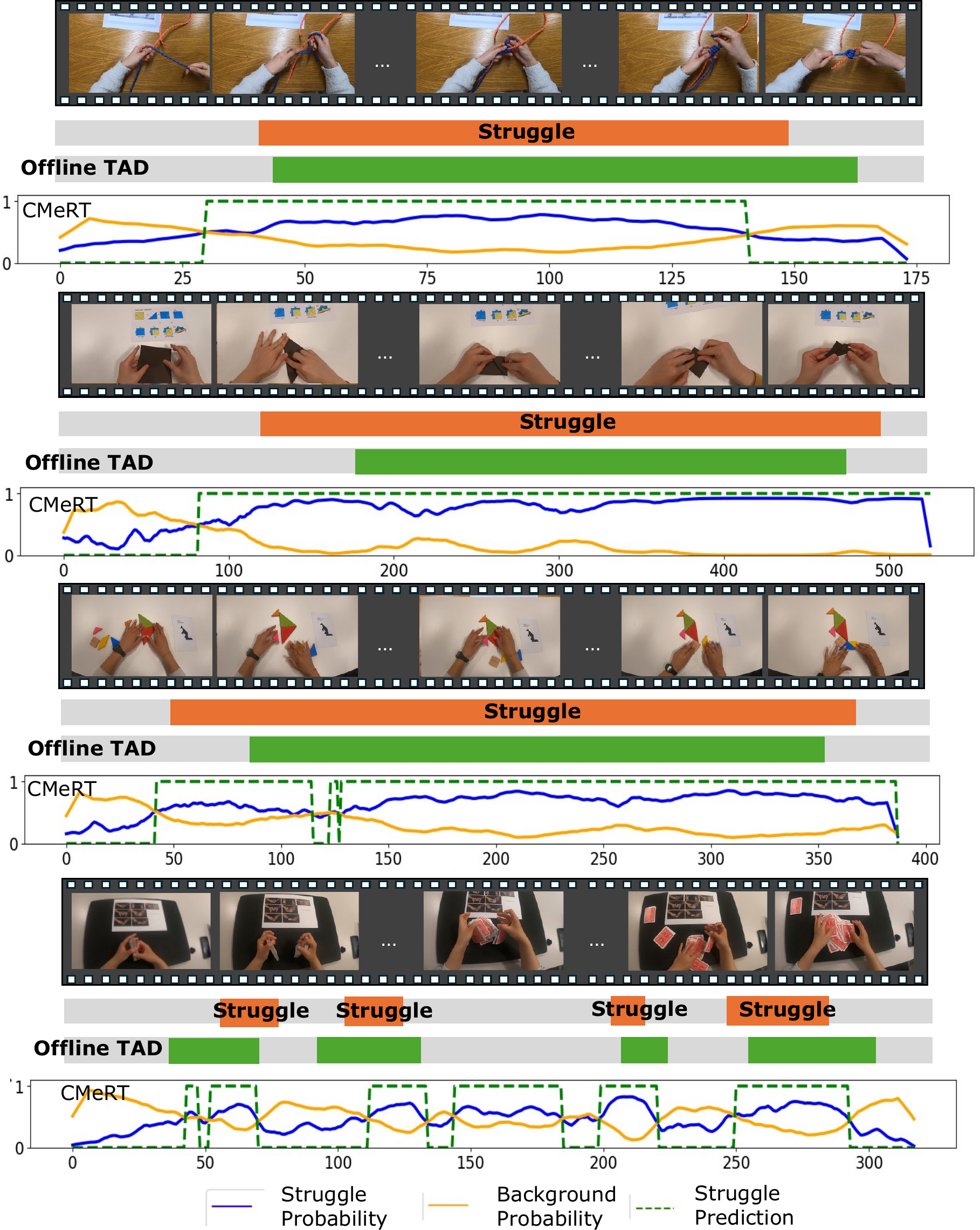}
% \vspace{-5mm}
\caption{Qualitative comparisons between the \textbf{Online Struggle Detection} with Model CMeRT~\cite{Pang_2025_CVPR} and the \textbf{offline Struggle Detection} in~\cite{fengevostruggle} using Model TriDet~\cite{10203543} (shown using green bars), across Activities (from top to bottom) Tying Knots, Origami, Tangram, and Shuffle Cards.}
% \vspace{-5mm}
\label{fig:online-offline-comparison}
\end{figure}

Finally, Figure~\ref{fig:online-offline-comparison} presents a qualitative comparison between the reformulated online struggle detection and the original offline approach in~\cite{fengevostruggle}. Both methods are able to detect struggle moments in the provided examples; however, the online model identifies them more promptly. In the first three examples, the online method captures the onset of struggle more accurately, and in some cases even slightly earlier than the annotated ground truth, whereas the offline method typically detects the onset with noticeable delay. On the other hand, the online approach can occasionally overfit, leading to false positives—as illustrated in the fourth example, where a spurious struggle detection appears mid-video.

% \begin{figure*}[ht]
% \centering
% \includegraphics[width=\linewidth]{wacv-2026-author-kit-template/figures/qualitative_results_struggle_ant-origami.png}
% % \vspace{-5mm}
% \caption{Qualitative Results on Struggle Anticipation Comparing Model LSTR~\cite{xu_long_2021} and Model CMeRT~\cite{Pang_2025_CVPR} on Activity Origami.}
% % \vspace{-5mm}
% \label{fig:struggle-anticipation-visualization-tyingknots}
% \end{figure*}

% \begin{figure*}[ht]
% \centering
% \includegraphics[width=\linewidth]{wacv-2026-author-kit-template/figures/qualitative_results_struggle_ant-tangram.png}
% % \vspace{-5mm}
% \caption{Qualitative Results on Struggle Anticipation Comparing Model LSTR~\cite{xu_long_2021} and Model CMeRT~\cite{Pang_2025_CVPR} on Activity Tangram.}
% % \vspace{-5mm}
% \label{fig:struggle-anticipation-visualization-tyingknots}
% \end{figure*}

% \begin{figure*}[ht]
% \centering
% \includegraphics[width=\linewidth]{wacv-2026-author-kit-template/figures/qualitative_results_struggle_ant-shufflecards.png}
% % \vspace{-5mm}
% \caption{Qualitative Results on Struggle Anticipation Comparing Model LSTR~\cite{xu_long_2021} and Model CMeRT~\cite{Pang_2025_CVPR} on Activity Shuffle Cards.}
% % \vspace{-5mm}
% \label{fig:struggle-anticipation-visualization-tyingknots}
% \end{figure*}

{
    \small
    \bibliographystyle{ieeenat_fullname}
    \bibliography{main}
}